\documentclass[10pt,twocolumn,letterpaper]{article}

\usepackage[pagenumbers]{cvpr}   

\usepackage{microtype}
\renewcommand{\paragraph}[1]{\vspace{.5em}\noindent\textbf{#1.}}
\usepackage{xspace}
\usepackage{multirow}
\usepackage{siunitx}
\usepackage{algorithm}
\usepackage{algpseudocode}
\usepackage{tikz}
\usetikzlibrary{calc}
\definecolor{streamcollect}{HTML}{F4DEDB}  
\definecolor{streamwm}{HTML}{ECEBE6}       
\definecolor{streampolicy}{HTML}{DFE6F0}   
\newcommand{\algmark}[1]{\tikz[remember picture,overlay]\coordinate (#1);}  
\newcommand{\algshade}[5]{\tikz[remember picture,overlay]\fill[#1,blend mode=multiply] let \p5 = (algL), \p1 = (#2),
  \p2 = (#3), \p3 = (#4), \p4 = (#5) in ({\x5-\leftmargin-0.2em}, {\y2+0.58*(\y1-\y2)})
  rectangle ({\x5+\linewidth+0.2em}, {\y3-0.42*(\y3-\y4)});}
\newcommand{\msh}{MS-Human-700\xspace}
\newcommand{\mfb}{MyoFullBody\xspace}

\newcommand{\dg}{\ensuremath{^\circ}\xspace}
\newcommand{\mm}{\,mm\xspace}
\newcommand{\cm}{\,cm\xspace}
\newcommand{\best}[1]{\textbf{#1}}

\newcommand{\R}{\mathbb{R}}

\definecolor{cvprblue}{rgb}{0.21,0.49,0.74}
\usepackage[breaklinks,colorlinks,allcolors=cvprblue]{hyperref}

\newif\ifarxiv \arxivtrue
\newcommand{\codeabs}{Code, the training recipe and videos are on the project page
  \url{https://muscle-supermimic.beanpow.top/}.}
\newcommand{\codeintro}{The code, the training recipe and videos are on the project page
  \url{https://muscle-supermimic.beanpow.top/}.}

\title{Fast Muscle-Driven Whole-Body Motion Tracking\\with a Differentiable World Model}

\author{Kaibo He$^1$ \qquad Maciej Bakowicz$^1$ \qquad Shihe Zhou$^2$ \qquad Yusen Feng$^3$\\[2pt]
Yanan Sui$^{2\dagger}$ \qquad Dhanush Radhakrishnan$^{1\dagger}$\\[2pt]
$^1$Clone Robotics \qquad $^2$Tsinghua University \qquad $^3$Peking University
}

\begin{document}
\twocolumn[{%
\renewcommand\twocolumn[1][]{#1}%
\maketitle
\begin{center}
  \captionsetup{type=figure}
  \includegraphics[width=\linewidth]{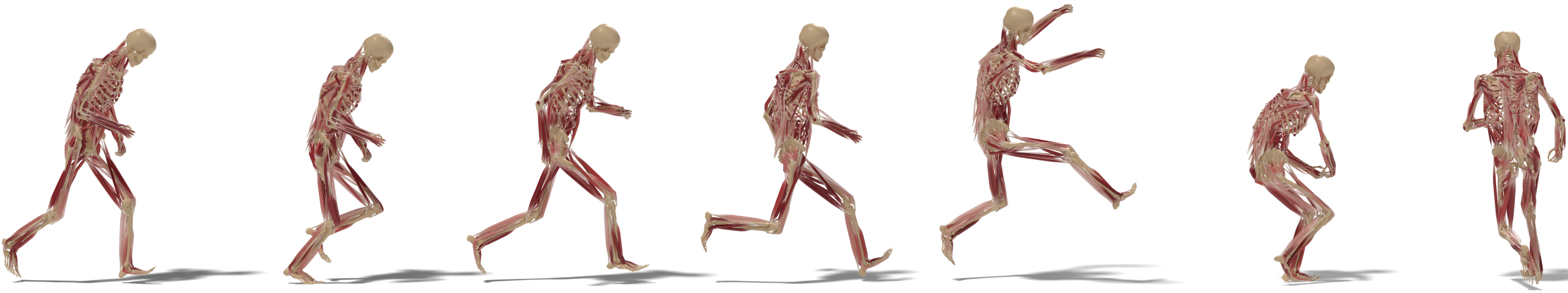}
  \captionof{figure}{\textbf{\msh tracking a LAFAN1 motion.} Seven instants of a 5-second training clip: walking,
  running into a leap, a crouch and a turn. The more a muscle is activated, the darker and more opaque it is
  drawn. Policy trained for one hour, with test-time refinement (\cref{sec:refine-method}).}
  \label{fig:teaser}
\end{center}
\vspace{0.5em}

}]
\begin{NoHyper}\renewcommand\thefootnote{}\footnotetext{\raggedright $^\dagger$Corresponding authors:
  {\tt ysui@tsinghua.edu.cn}, {\tt dhanush@clonerobotics.com}}\addtocounter{footnote}{-1}\end{NoHyper}
\begin{abstract}
Tracking motion capture with a whole-body musculoskeletal model means coordinating hundreds of muscles through contact,
and current controllers take hours to days of GPU time to train, so every tuning decision is expensive. SuperMimic trains a tracking policy by backpropagating through the rollouts of a learned world model. We extend it to
\msh (700 muscles) and \mfb (354 muscles), with MuJoCo's activation dynamics inside the world model. To make its training fast, our recipe takes the speedups developed for the Unitree G1, replaces three of their
settings that fail on muscles, and adds four changes of our own: observations without muscle geometry, data collection
concurrent with the updates, Beta exploration noise that keeps every excitation in $[0,1]$, and 50~Hz control. On one
RTX~5090 it trains \msh in \textbf{6.3 minutes} to the held-out return that SuperMimic's published hyperparameters reach
in 3.9 hours, 37$\times$ sooner, and in 20 minutes to lower tracking errors than theirs after 10 hours. It trains \mfb
34$\times$ sooner without retuning.
Ablations show what is specific to muscles: a higher policy learning rate stalls half of the
runs, and clipped Gaussian noise adds a tonic co-contraction that Beta noise avoids. Finally, refining the policy's
actions through the world model at test time lowers the position error by a further 26\% with no held-out clip failing, and
reaches the same return after 5.1 instead of 6.3 minutes of training. \codeabs
\end{abstract}

\section{Introduction}
\label{sec:intro}

Human musculoskeletal models are where motor control, biomechanics and embodied AI meet. A whole-body
model such as \msh~\cite{zuo2024mshuman700} or MyoSuite's \mfb~\cite{caggiano2022myosuite,li2026musclemimic} has
hundreds of Hill-type muscles~\cite{hill1938heat} pulling on tendons that wrap around bones, nonlinear activation dynamics, redundant
actuation at every joint and no direct torque control. Driving such a model to reproduce human motion capture is the task
behind simulated biomechanics~\cite{delp2007opensim,rajagopal2016fullbody}, assistive-device design~\cite{luo2024exoskeleton},
physically plausible animation of muscle-driven characters~\cite{lee2019scalable}, and
the study of human motor learning itself. It is also hard: a tracking policy has to coordinate 354 to 700 excitations
so that the model follows a reference, through contacts, for clips it has never seen.

Model-free reinforcement learning solves the task at a cost. MuscleMimic~\cite{li2026musclemimic} trains a whole-body
\mfb tracker on the KIT motions of AMASS with PPO~\cite{schulman2017ppo} at $1.3 \times 10^4$ environment steps per second, and its released
tracker took 3.2 billion steps, about three days on an H100. MS-Emulator~\cite{wei2026msemulator} trains \msh with PPO and reports 7 hours on one RTX~5090 to emulate one
running trajectory. KINESIS~\cite{simos2025kinesis}, DEP-RL
\cite{schumacher2023deprl} and MuscleVAE~\cite{feng2023musclevae} report hours of training as well, on legs or on smaller
musculoskeletal models. Hours per run make a recipe hard to tune: with one GPU, a two-seed comparison of one hyperparameter takes a day.

For rigid humanoids the picture has changed. SuperMimic~\cite{liao2026supermimic} trains a whole-body tracker for the
Unitree G1 by backpropagating the tracking reward through imagined rollouts of one learned world model of the robot, and
on a 40-clip LAFAN1 benchmark reaches in 1.2 hours a return that PPO in the simulator does not reach in 18 hours. Systems work on top of
it, InstantMimic~\cite{choi2026instantmimic}, InstantTrack~\cite{instanttrack2026} and BlinkMimic~\cite{blinkmimic2026},
brought the same benchmark to about three minutes on one RTX~4090. None of it has been shown on muscles, where the action is an
excitation in $[0,1]^{700}$ with its own dynamics and the world model has to predict forces that come from muscle
geometry. There, three of the settings that make the G1 recipe fast break the training (\cref{sec:recipe}).

This paper makes world-model tracking work on whole-body muscle models and makes it fast. We first extend SuperMimic to
\msh and \mfb. MuscleMimic's tracking MDP is written as analytic functions of the state, so that the
simulator and the imagined rollouts evaluate the same observation, reward and terminations. The world model integrates
the muscle activations with MuJoCo's own substep rule, and a differentiable torch replica of MuJoCo's spatial tendons
and Hill muscles gives muscle length, velocity and force on imagined states.
With SuperMimic's published hyperparameters, this extension, our \emph{baseline}, already
trains \msh to a
normalized return of 0.83 on MuscleMimic's 108 held-out clips in 3.9 hours and \mfb in 6.0 hours, on one RTX~5090.

We then make the training fast. We adopt the G1 speedups, replace the three that do not transfer
(unclamped bootstrapped values without a world-model warm-up, a $4\times$ early policy learning rate, and
evaluation-guided resets on 5~s clips), and add four changes of our own. The policy observes no muscle
geometry, so the replica, the costliest part of an imagined step, leaves the training loop. The collection step, captured as one CUDA graph as in
BlinkMimic, runs on a second CUDA stream concurrently with the updates, so that 1024 environments suffice. The
exploration noise is drawn from a Beta distribution centred on the policy's excitation, because a clipped Gaussian adds
a tonic co-contraction to every muscle the policy keeps silent. The policy acts at 50~Hz with every step-counted
setting rescaled. We measure every step (\cref{fig:ladder}). In total, the time to 0.83
goes from 3.9 hours to 6.3 minutes, the time per iteration from 2.3 to 0.14~s, and the
20-minute policies track the 108 held-out clips with a mean per-joint position error (MPJPE) of 18\mm and a joint-angle
error of 3.4\dg.

Minutes per run make systematic ablation affordable, and we report tests of the recipe's
remaining knobs. Some outcomes are specific to muscles. A higher policy learning rate, which helps on the G1, stalls half
of the runs. The exploration noise distribution decides between 8.7 and 13.7 minutes to 0.83, or not reaching it. DEP-RL's
embodied exploration~\cite{schumacher2023deprl}, designed for muscles, does not help, and halving the control rate is
only free if the exploration noise is rescaled with it.

Finally, the policy does not fully exploit the learned world model. Moving the policy's action
at test time by a few gradient-ascent steps on the imagined return lowers the MPJPE by a further 26\% and the
global position error by 53\%, and tracks every held-out clip. Part of the remaining
tracking error is the policy not reaching the world model's optimum, a gap that regression onto such actions in
training does not close.

Our contributions are: (1) SuperMimic on two whole-body muscle models, with MuJoCo's activation dynamics inside the
world model and one analytic MDP on both sides; (2) a training recipe that reaches the baseline's return
37$\times$ sooner, with every step measured and the recipe's remaining knobs ablated on \msh; (3) tracking accuracy on MuscleMimic's KIT split measured with the metrics
of prior work, with MuscleMimic's released tracker evaluated on the same clips and metrics (after 20 minutes instead of
its three days, ours fail fewer clips, with lower joint errors but higher position errors); and (4) test-time action
refinement through the world model, which tracks every held-out clip and lowers the MPJPE by a further 26\% on \msh and
below MuscleMimic's on \mfb. \codeintro

\section{Related work}
\label{sec:related}

\paragraph{Musculoskeletal control}
MyoSuite~\cite{caggiano2022myosuite} brought contact-rich musculoskeletal models to MuJoCo~\cite{todorov2012mujoco}.
On them MyoDex~\cite{caggiano2023myodex} learns a multi-task prior for dexterous hand manipulation and
SAR~\cite{berg2023sar} transfers muscle synergies learned on simpler tasks. \msh~\cite{zuo2024mshuman700} is a
whole-body model with 700 muscles, on which DynSyn~\cite{he2024dynsyn} learns synergistic low-dimensional actions and
MPC$^2$~\cite{wei2025mpc2} plans. DEP-RL~\cite{schumacher2023deprl}, Lattice~\cite{chiappa2023lattice} and BioSynGrasp~\cite{liu2026biosyngrasp} explore overactuated systems,
with a self-organizing controller, with noise injected into the policy's latent state and with correlated muscle perturbations. Motion tracking on muscles has been learned model-free. Lee \etal~\cite{lee2019scalable} imitate motion capture with
a full-body muscle-actuated character, and KINESIS~\cite{simos2025kinesis} imitates KIT locomotion on a leg model with
PPO. MuscleMimic~\cite{li2026musclemimic} scales PPO to full-body \mfb tracking on 972 KIT clips and reports 92.6\%
success on the 108 held-out clips after billions of steps. TERRA~\cite{simos2026terra} retargets locomotion onto
reconstructed terrain and trains one \mfb policy on 9.4 hours of it with PPO on MJX-Warp, in about 20 hours on four A100
GPUs. MuscleVAE~\cite{feng2023musclevae} instead trains a model-based controller of a muscle-actuated character. MS-Emulator~\cite{wei2026msemulator} is the closest to ours: whole-body motion emulation on \msh with PPO on
GPU-parallel MuJoCo-Warp, an adversarial tracking reward and value-guided flow exploration \cite{wei2026qflex}. It
reports 7 hours on one RTX~5090 to emulate one running trajectory at a mean joint-angle error of at most 2\dg over all
joints, the same for walking, and at most 7\dg and about 6\cm on highly dynamic AMASS motions. We use the same whole-body musculoskeletal model,
simulator and joint-angle definition (\cref{sec:accuracy}), reach 1.9\dg on 108 held-out clips with one policy
trained for 20 minutes, and show what in the fast G1 recipes fails on muscles, which none of these works examines.

\paragraph{Tracking through learned dynamics}
SuperTrack~\cite{fussell2021supertrack} and ControlVAE~\cite{yao2022controlvae} train controllers for a simulated character by backpropagating through a
learned world model, and DiffMimic~\cite{ren2023diffmimic} and SHAC~\cite{xu2022shac} through a differentiable simulator, the latter with a critic bootstrap.
SuperMimic~\cite{liao2026supermimic} scales the former to whole-body humanoid control on a real robot: one world model of
the robot, analytic rewards and soft terminations on imagined states, a critic that bootstraps the imagined return as in Dreamer~\cite{hafner2020dreamer} and MAAC~\cite{clavera2020maac}, and
thousands of parallel environments. Open implementations made it fast: InstantMimic~\cite{choi2026instantmimic}
profiled and fused the training loop, InstantTrack~\cite{instanttrack2026} and BlinkMimic~\cite{blinkmimic2026} removed
host synchronization, captured the updates and the collection step as CUDA graphs and added Muon~\cite{jordan2024muon,liu2025muon},
a semi-implicit world model, mirror augmentation and evaluation-guided resets, reaching 0.82 on the G1 in minutes. NDM Gym~\cite{dou2026ndmgym} benchmarks SuperTrack and SuperMimic side by side and
extends them to a muscle-actuated character. $\nabla$Mimic~\cite{lu2026nablamimic} reaches the same 0.82 in 20 minutes
from the exact gradients of a differentiable G1 simulator, clipped per sample and regressed into the policy as action
targets. We build on the learned-model recipes and report, on 700 muscles, which of their settings transfer, which have
to change and which do not matter (\cref{sec:recipe,sec:ablations}).

\paragraph{Humanoid motion tracking}
DeepMimic~\cite{peng2018deepmimic} set the task and the reward form, and PHC~\cite{luo2023phc} tracks large motion-capture datasets with simulated avatars; BeyondMimic~\cite{liao2025beyondmimic},
HDMI~\cite{weng2025hdmi} and SONIC~\cite{luo2026sonic} track human motion on a real G1 with model-free RL in massively parallel GPU
simulation~\cite{rudin2022learning,mittal2025isaaclab,zakka2026mjlab}. Our task follows MuscleMimic's formulation \cite{li2026musclemimic}
on KIT motions of AMASS~\cite{mahmood2019amass,mandery2016kit} retargeted with GMR~\cite{araujo2025gmr}, so that the
numbers are comparable.

\paragraph{Test-time action refinement}
POPLIN~\cite{wang2020poplin} and TD-MPC2~\cite{hansen2024tdmpc2} plan with a learned model from a policy's proposals by sampling.
Qflex~\cite{wei2026qflex} refines sampled actions by gradient ascent on a learned $Q$ and trains a flow to produce them, for
exploration in high-dimensional control. $\nabla$Mimic's action targets, after PODS~\cite{mora2021pods}, take the same
step through an exact simulator, in training. We use the refinement alone, at test time, with the world model's
imagined return in place of $Q$, as a probe of the gap between the policy and the model (\cref{sec:refine}).

\section{Method}
\label{sec:method}

\begin{figure*}[t]
  \centering
  \includegraphics[width=\linewidth]{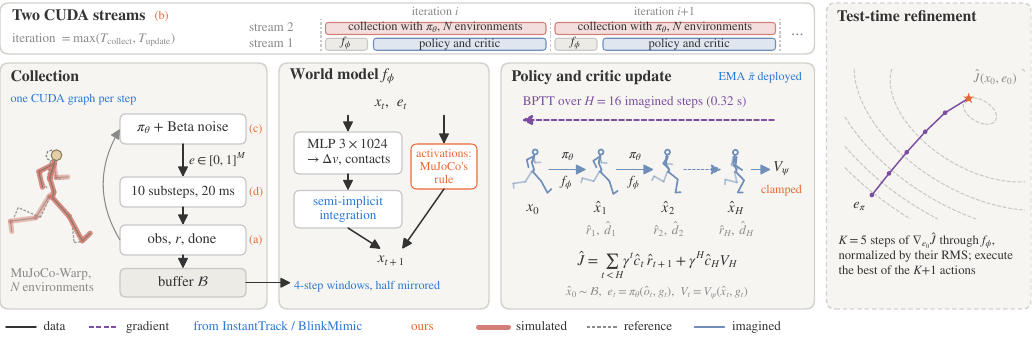}
  \caption{\textbf{Overview of the training loop and the test-time refinement.} In \emph{collection}, $N$ parallel environments act with
  the policy $\pi_\theta$ and exploration noise, each control step one CUDA graph. Their transitions fill a
  buffer. The \emph{world model} $f_\phi$ predicts velocity increments and contacts and integrates the positions
  semi-implicitly and the activations by MuJoCo's own rule. In the \emph{policy and critic update}, the tracking return $\hat J$ (\cref{eq:imag}) of an $H$-step rollout of $\pi_\theta$ through $f_\phi$ from start states of $\mathcal B$ is backpropagated into $\pi_\theta$. The collection runs on a second CUDA stream concurrently with the two updates. At test time, the
  policy's action is refined by gradient ascent on $J$ through $f_\phi$ (\cref{sec:refine-method}). Blue marks what comes from InstantTrack and BlinkMimic and orange what is ours, with (a)--(d) the changes of \cref{sec:recipe}.}
  \label{fig:overview}
\end{figure*}

We train a tracking policy for a whole-body musculoskeletal model by backpropagating the tracking reward through a learned
world model, and make this training fast. \Cref{sec:task} states the two whole-body musculoskeletal models and the task, \cref{sec:learning} the learning
through the world model with what muscles add to it, \cref{sec:recipe} the recipe that makes it fast, and
\cref{sec:refine-method} the refinement of the policy's actions through the same model at test time.

\subsection{Whole-body musculoskeletal models and task}
\label{sec:task}

\paragraph{Whole-body musculoskeletal models}
\msh~\cite{zuo2024mshuman700} has 79 joints besides the root, 37 of them independent and 42 tied to these by polynomial
equality constraints, and 700 Hill-type muscles, each pulling a spatial tendon that wraps around spheres and cylinders.
We lock its hands (the wrist joints welded in the neutral pose, the 6 muscles that cross only them removed), which leaves
71 joints, 33 of them independent, and 694 muscles. \mfb is MyoSuite's full-body model as
MuscleMimic~\cite{li2026musclemimic} configures it for tracking, without fingers: 82 joints, 31 of them independent,
354 muscles and 75 self-contact pairs. Both are simulated in mjlab~\cite{zakka2026mjlab} on
MuJoCo-Warp~\cite{todorov2012mujoco,mujocowarp} with MuscleMimic's solver settings, thousands of environments in parallel on one GPU
(\cref{sec:supp-task}).

\paragraph{Muscles}
The state is $x = (p, R, v, \omega, q, \dot q, a)$: the root's position, orientation and linear and angular velocity,
the joint positions and velocities, and the activations $a \in [0,1]^M$ of the $M$ muscles. The policy sets the
excitations $e \in [0,1]^M$ once per control step $\Delta t$, and the activations
follow MuJoCo's first-order dynamics~\cite{millard2013flexing},
\begin{equation}
  \dot a = \frac{e - a}{\tau(e, a)}, \qquad a_{t+1} = A(a_t, e_t),
  \label{eq:act}
\end{equation}
with $\tau$ the activation time constant when $e > a$ and the slower deactivation one otherwise, both scaled with $a$,
and $A$ the map over one control step: $n$ explicit Euler steps of length $\Delta t/n$, the physics substeps, with
$e_t$ held fixed.
Muscle $m$ pulls along its tendon path, of length $\ell_m = L_m(q)$, with the Hill force~\cite{hill1938heat}
\begin{equation}
  F_m = F^{\max}_m \big[a_m\, f_L(\tilde\ell_m)\, f_V(\dot{\tilde\ell}_m) + f_P(\tilde\ell_m)\big],
  \label{eq:hill}
\end{equation}
where $\tilde\ell_m$ is the normalized length and $f_L$, $f_V$ and $f_P$ are the active force-length, force-velocity and
passive force curves.

\paragraph{Tracking task}
We use MuscleMimic's tracking MDP unchanged, so that the numbers are comparable (\cref{sec:supp-task}). The policy
observes the skeletal state and the last excitations, each muscle's $\ell_m$, $\dot\ell_m$ and $F_m$, four foot-contact forces and 17 tracked
sites relative to the root. The goal $g_t$ is the reference's sites at the current and four future frames 0.2~s apart,
the root deltas and the clip phase. With the reference frame $x^{\mathrm{ref}}_t$, the reward is a weighted sum of
Gaussian kernels on the errors $\varepsilon_i$ of the joints, the root and the sites,
\begin{equation}
  r(x_t, x^{\mathrm{ref}}_t) = \textstyle\sum_i w_i \exp\!\big(-k_i\, \varepsilon_i(x_t, x^{\mathrm{ref}}_t)\big),
  \label{eq:reward}
\end{equation}
and an episode terminates when the sites, the root position or the root orientation deviate from the reference by more
than 1~m, 1~m or 1~rad. The motions are MuscleMimic's GMR-retargeted KIT clips of
AMASS~\cite{mahmood2019amass,mandery2016kit,araujo2025gmr} in the KINESIS split, 972 training clips (1.5~h) and 108 held-out
ones. They drive \mfb as they are, and we transfer them kinematically to \msh (\cref{sec:supp-task}).

\subsection{Learning through a differentiable world model}
\label{sec:learning}

\paragraph{Training loop}
We learn a world model $f_\phi$ from simulator transitions and train the policy $\pi_\theta$ by backpropagating
the analytic tracking reward through rollouts of $f_\phi$ (\cref{fig:overview}), the scheme of
SuperMimic~\cite{liao2026supermimic}. Our implementation starts from its open re-implementation~\cite{zhang2026supermimicvibed}. Each
iteration collects transitions of $N$ parallel environments into a replay buffer $\mathcal B$, fits $f_\phi$ on $k$-step
windows of $\mathcal B$ and updates the policy and a critic $V_\psi$ on imagined rollouts from states of $\mathcal B$
(\cref{alg:train}). The policy is deterministic, $e_t = \pi_\theta(o_t, g_t)$ with the observation
$o_t = o(x_t)$. The simulator executes $e_t$ with exploration noise, and $\mathcal B$ stores the noisy excitation. With the termination
$d_t \in \{0,1\}$ of $x_t$ and the survival weight $c_0 = 1$, $c_{t+1} = c_t(1 - d_{t+1})$, the objective is
$J(\theta) = \mathbb E\big[\sum_{t \ge 0} \gamma^t c_t\, r_{t+1}\big]$, $r_{t+1} = r(x_{t+1}, x^{\mathrm{ref}}_{t+1})$.

\paragraph{World model}
$f_\phi$ predicts the next state, $\hat x_{t+1} = f_\phi(\hat x_t, e_t)$, with a hat marking a quantity imagined through $f_\phi$ rather than simulated. The observation, reward and terminations of
an imagined state are the analytic functions of the simulator side, $\hat o_t = o(\hat x_t)$ and
$\hat r_{t+1} = r(\hat x_{t+1}, x^{\mathrm{ref}}_{t+1})$, with the terminations softened into
$\hat d_t = 1 - \prod_i \big(1 - \mathrm{sigmoid}((\rho_i - \bar\rho_i)/0.05)\big)$ over the deviations
$\rho_i(\hat x_t)$ of the sites, the root position and the root orientation and their thresholds $\bar\rho_i$.
$f_\phi$ advances the activations by \cref{eq:act}
exactly as the simulator does, which matches them to float precision. An MLP on heading-invariant features of
$\hat x_t$ and on $\hat a_t$ and $\hat a_{t+1}$ learns the rest, the increments of the velocities $\nu = (v, \omega,
\dot q)$ and the foot contacts. The positions $s = (p, R, q)$ follow from the old and new velocities as in the $n$
semi-implicit Euler substeps~\cite{blinkmimic2026}, plus a predicted correction,
\begin{equation}
  \nu_{t+1} = \nu_t + \Delta\nu_\phi, \qquad
  s_{t+1} = s_t + \Delta t \big[(1 - w)\,\nu_t + w\,\nu_{t+1}\big] + \delta s_\phi,
  \label{eq:wm}
\end{equation}
with $w = (n+1)/(2n)$ the weight of the new velocity over the substeps, the root terms in its heading frame and the
orientation updated by a rotation vector. $f_\phi$ is fitted on $k$-step windows of $\mathcal B$, rolled out from their
first state $\hat x_0 = x_0$ on the logged excitations, half of each batch mirrored left to right (\cref{sec:supp-mirror}):
\begin{equation}
  L(\phi) = \mathbb E_{\mathcal B} \sum_{j=1}^{k} \Big[\tfrac{1}{D}\big\|(\hat x_j \ominus x_j) \oslash \sigma\big\|^2 + \tfrac{\beta}{C}\, \|\hat\kappa_j - \kappa_j\|^2\Big],
  \label{eq:wmloss}
\end{equation}
with $\hat\kappa_j$ the predicted foot contacts, $\ominus$ the state difference without the activations, $\oslash$
elementwise division, $\sigma$ the standard deviation of each component's increments, and $D$ and $C$ the numbers
of state and contact components. The reward, observation and terminations are analytic and
add no terms.

\paragraph{Policy and critic}
With $f_\phi$ held fixed, the policy maximizes by backpropagation through time the return of $H$-step imagined rollouts
from states $\hat x_0$ of $\mathcal B$, with $e_t = \pi_\theta(\hat o_t, g_t)$,
\begin{equation}
  \hat J(\theta) = \mathbb E\Big[\sum_{t<H} \gamma^t \hat c_t\, \hat r_{t+1} + \gamma^H \hat c_H\, V_\psi(\hat x_H, g_H)\Big],
  \label{eq:imag}
\end{equation}
with $\hat c_t$ the survival weight on $\hat d_t$. The critic, which also sees privileged reference errors, is fitted to the
$\lambda$-returns of the same rollouts, computed once per rollout and held fixed while it is fitted:
\begin{equation}
  \begin{gathered}
  L_V(\psi) = \mathbb E \textstyle\sum_{t<H} \big(V_t - \hat R_t\big)^2, \qquad \hat R_H = V_H, \\
  \hat R_t = \hat r_{t+1} + \gamma (1 - \hat d_{t+1}) \big[(1-\lambda) V_{t+1} + \lambda \hat R_{t+1}\big],
  \end{gathered}
  \label{eq:critic}
\end{equation}
with $V_t = V_\psi(\hat x_t, g_t)$. The policy maps the observation and an encoding of the goal to
$e = \mathrm{sigmoid}(\cdot)$, and a Polyak average $\bar\pi$ of it is deployed~\cite{instanttrack2026}.

\paragraph{Differentiable muscles}
On imagined states the observation needs each muscle's $\ell_m$, $\dot\ell_m$ and $F_m$, which the simulator computes from
the tendon paths and \cref{eq:hill}. We implemented a torch replica of MuJoCo's spatial tendons $L_m(q)$, with their
sphere and cylinder wraps, and of \cref{eq:hill}, batched over environments and muscles and differentiable in $x$. It equals
MuJoCo's lengths, velocities and forces to $10^{-6}$ or better (\cref{sec:supp-replica}). The baseline uses it for
MuscleMimic's full observation (\cref{sec:speed,sec:mfb}). The fast recipe drops that observation and with it the
replica (\cref{sec:recipe}, (a)), which any observation or reward on muscle length, velocity or force would need again.

\subsection{The fast recipe}
\label{sec:recipe}

The baseline, our extension with SuperMimic's published hyperparameters, reaches a held-out return of
0.83 on \msh after 3.9 hours on one
RTX~5090 (2.3~s per iteration). The recipe below reaches it in 6.3 minutes.
\Cref{tab:hparams} lists both settings.

\paragraph{Adopted from the G1 recipes}
We take from InstantTrack~\cite{instanttrack2026} and BlinkMimic~\cite{blinkmimic2026}, unchanged, the Polyak-averaged
policy, Muon~\cite{jordan2024muon,liu2025muon} on the hidden weight matrices, the semi-implicit world model of
\cref{eq:wm} with mirror augmentation and resets focused just before recent failures. The world-model update, the policy
and critic update and the collection step are captured as CUDA graphs, the last through a thin wrapper of the simulator,
with no host synchronization (\cref{sec:supp-hparams,sec:supp-collector}).

\paragraph{Changed for 700 muscles}
Three of BlinkMimic's settings fail on \msh (\cref{sec:supp-firstruns}). (i) \emph{Unclamped bootstrapped values with
no world-model warm-up}: the critic diverges from the first policy updates, its imagined return reaching more than ten times
the attainable maximum. We train $f_\phi$ alone for 200 iterations and clamp the bootstrap to
$V_\psi(x_H) \le r_{\max}/(1-\gamma)$, with $r_{\max}$ the reward of a perfect step. (ii) \emph{A $4\times$ policy learning
rate early on}: the policy is ahead at first, then degrades and does not recover (\cref{sec:ablations}). We keep $1\times$.
(iii) \emph{Evaluation-guided resets}, which probe every clip with the deployed policy and focus the resets before its
failures: on 5~s clips every probe fails within 3~s of its start, so the focus covers whole clips, and the probing costs
half of the training time. We turn them off.

\paragraph{Ours}
(a) \emph{No muscle geometry in the observation.} $\ell_m$, $\dot\ell_m$ and $F_m$ are functions of the joint state and
activations the policy still sees. Dropping them removes the tendon paths, the costliest part of an imagined step, from
the policy update.
(b) \emph{The collection concurrent with the updates.} The graphed collection step covers a whole simulator step of
all environments, from the analytic observation, the policy and its noise and MuJoCo-Warp's substeps to the rewards,
terminations, buffer write and masked resets. It is replayed on a second CUDA stream with its own copy of the policy,
while the updates read the buffer as it stood at the start of the iteration (\cref{sec:supp-collector}). An iteration then
takes $\max(T_{\mathrm{collect}}, T_{\mathrm{update}})$ instead of their sum, and 1024 environments suffice.
We also turn off rematerialization, keeping the intermediate tensors of the imagined steps for the backward pass instead
of recomputing them, and compile the simulator-side kinematics precompute.
(c) \emph{Exploration noise in the excitation box.} The baseline adds SuperMimic's Gaussian noise $e + \sigma\epsilon$,
$\epsilon \sim \mathcal N(0,1)$, and clips it to $[0,1]$~\cite{fujita2018clipped}, which is not mean-preserving near the bounds: for $e \ll 1$ the executed excitation has the mean
\begin{equation}
  \mathbb E\big[\mathrm{clip}(e + \sigma\epsilon)\big] = e\,\Phi(e/\sigma) + \sigma\,\Phi'(e/\sigma) \;\ge\; \sigma/\sqrt{2\pi},
  \label{eq:clipbias}
\end{equation}
with $\Phi$ the standard normal distribution function: 0.04 for $\sigma = 0.1$ however silent the muscle. A trained policy keeps a quarter of its excitations below 0.05, and the
noise executes them at 0.050 on average where the policy asks for 0.0185 (\cref{sec:supp-beta}): a tonic co-contraction
that the deterministic policy never produces and the world model has to fit. We draw instead, as Beta policies do~\cite{chou2017beta}, from
$\mathrm{Beta}(\kappa e, \kappa(1-e))$ with $\kappa = 1/(4\sigma^2) - 1$, whose mean is $e$ and whose standard deviation
$\sqrt{e(1-e)/(\kappa+1)}$ is $\sigma$ at $e = 1/2$ and shrinks towards the bounds.
(d) \emph{50~Hz control.} The policy acts every 20~ms, so an imagined horizon of the same duration has half the steps.
Every setting counted in steps keeps its duration, every per-step factor its rate~\cite{tallec2019discretization}, and the exploration noise the
variance of its integral over time:
\begin{equation}
  n' = n\,\Delta t/\Delta t', \quad \gamma' = \gamma^{\Delta t'/\Delta t}, \quad
  \sigma' = \sigma \sqrt{\Delta t/\Delta t'},
  \label{eq:rate}
\end{equation}
applied to the horizon $H$, the goal-frame spacing, the buffer, the world-model windows and the simulator steps per
iteration ($n$), to $\gamma$ and $\lambda$, and to the noise ($\sigma' = 0.0707$).

\subsection{Test-time action refinement through the world model}
\label{sec:refine-method}

The policy is trained to maximize \cref{eq:imag} under the world model, but a few gradient steps of a network on a batch
need not reach the optimum of every state. At test time, from the simulator's state $x_0$, we move the deployed policy's
action $e_\pi$ by $K$ normalized gradient-ascent steps on the imagined return of \cref{eq:imag} with $e_0$ as its first
action and the deployed policy from the second step on, exactly as the policy update computes it:
\begin{equation}
  e^{(k+1)} = \mathrm{clip}\Big(e^{(k)} + \eta \, \frac{\nabla^{(k)}}{\mathrm{RMS}(\nabla^{(k)})}\Big),
  \label{eq:refine}
\end{equation}
with $\nabla^{(k)} = \nabla_{e_0} \hat J(x_0, e_0)\big|_{e_0 = e^{(k)}}$, and execute the one of the $K+1$ iterates (the policy's own action included) with the highest imagined return. The
$K+1$ imagined rollouts and $K$ gradients of one control step are captured as one CUDA graph, which makes the refinement
cheap to evaluate on all held-out clips at once (\cref{tab:refine-seeds}), though not real time. We use it as a probe of the model (\cref{sec:refine}).

\section{Experiments}
\label{sec:exp}

\paragraph{Setup}
All training runs use one NVIDIA RTX~5090 (32~GB), PyTorch~2.14 with bf16 MLPs, mjlab~1.6 and
MuJoCo-Warp. Unless stated otherwise the whole-body musculoskeletal model is \msh with locked hands, so the policy drives 694 of its 700 muscles
(\cref{sec:task}), and the data MuscleMimic's KIT split.
Training times are wall-clock with warm compilation caches, which we save and reload. On a fresh machine, the compiled
pieces and CUDA graphs cost about 9 minutes before the first policy update (\cref{sec:supp-compile}). The first runs
(the first four steps of \cref{fig:ladder}) include a few minutes of compilation and are upper bounds.

\paragraph{Evaluation protocol}
As in SuperMimic, every held-out clip plays once from its first frame with the deterministic deployed
policy and no exploration noise. A clip \emph{fails} if it triggers a termination. The \emph{normalized return} is the
reward sum divided by its maximum attainable value, averaged over the 108 clips. The \emph{time to 0.83} is the training
time of the first checkpoint (saved every 250 iterations) whose held-out return is at or above 0.83, the value
the baseline reaches after 3.9 hours on \msh (\cref{fig:ladder}), close to SuperMimic's own 0.82
criterion on the G1. Tracking errors are averaged over every control step before a clip's end or failure,
against its reference frame. They are the joint-angle MAE and RMSE over the independent joints and the MAE over all
joints (MS-Emulator's definition), the MPJPE and body RMSE of the tracked sites relative to the pelvis, the global MPJPE
and the root orientation error, all defined in \cref{sec:supp-metrics}. Every
number is on the 108 held-out clips unless the training clips are named.

\subsection{Time to a held-out return of 0.83}
\label{sec:speed}

\begin{figure}[t]
  \centering
  \includegraphics[width=\linewidth]{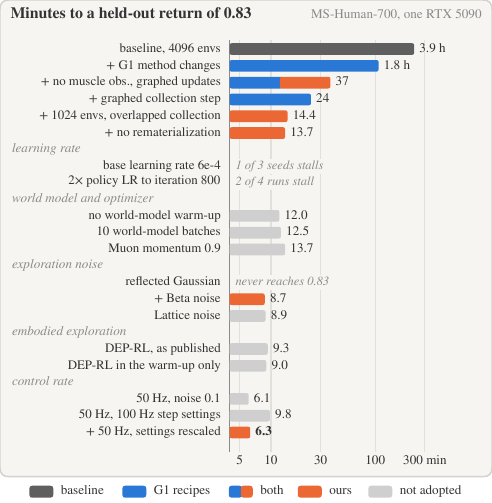}
  \caption{\textbf{Training time to a held-out return of 0.83 on \msh.} One RTX~5090, log scale. Each step is added
  to the ones above and coloured by where it comes from: the baseline (our extension with SuperMimic's published hyperparameters), the
  G1 recipes of InstantTrack and
  BlinkMimic (three settings changed for muscles), both, or ours. Grey rows are a selection of the alternatives tried at
  that point and not adopted. Every bar is the mean of three seeds, listed in \cref{tab:ladder,tab:ablation}.}
  \label{fig:ladder}
\end{figure}

\begin{figure}[t]
  \centering
  \includegraphics[width=\linewidth]{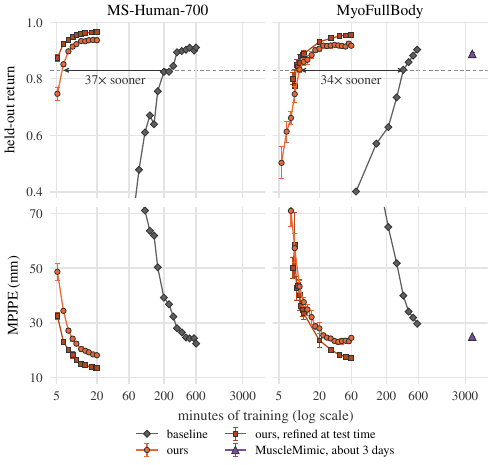}
  \caption{\textbf{Held-out return and MPJPE against training time.} Return (top) and MPJPE (bottom) on both whole-body musculoskeletal models: the
  baseline (one run), our recipe and the same checkpoints refined at test time. Mean of three seeds, error bars one
  standard deviation over the seeds. Arrows: the mean time to the
  first evaluation at or above 0.83 (grey line), for the \msh baseline over its three seeds (\cref{tab:ladder}). Triangle: MuscleMimic's released \mfb checkpoint, evaluated by us with
  the same protocol (\cref{tab:accuracy}), at its training time on an H100 (3.2 billion steps).}
  \label{fig:curves}
\end{figure}

\Cref{fig:ladder} adds the changes of \cref{sec:recipe} one group at a time, and \cref{tab:ladder} lists each step's
iterations and time per iteration. The G1 method changes, with the three muscle-specific replacements, halve the
iterations to 0.83. The systems changes then cut the time per iteration from 1.89 to 0.22~s with a nearly unchanged
learning curve. Beta exploration noise is a learning change: it reaches 0.83 a third sooner and tracks every clip
earlier (\cref{sec:ablations}). 50~Hz control halves the policy steps per simulated second and the imagined horizon in steps, and
with the noise rescaled it reaches 0.83 after 6.3~minutes, 37$\times$ sooner than the baseline. After 20
minutes its return exceeds the baseline's after 10 hours (\cref{fig:curves}).

\subsection{Ablations on \msh}
\label{sec:ablations}

The grey rows of \cref{fig:ladder} report the ablations, each one change to the last step above it.

\paragraph{Learning rate}
BlinkMimic's early $4\times$ policy learning rate wrecks the policy (\cref{sec:recipe}). A milder $2\times$ phase once reached
0.83 one checkpoint earlier, but stalled half of the replicate runs at a held-out return of 0.3--0.4 (imagined return
0.6, a third of the training episodes failing). On the same seeds $1\times$ stalled none and was not slower. A higher base rate ($6 \times 10^{-4}$ for all three optimizers) stalls a seed, and the others stay below the reference's best return.
Whatever the fast phase does persists after the rate has decayed: the stalled runs are level with the healthy ones at
iteration 1000 and never climb. A likely cause is that with nearly 700 excitations and a critic bootstrapping an imagined return, the policy can
move into states the model and critic have not seen faster than they can follow.

\paragraph{Exploration noise}
The three noises that keep the excitation in $[0,1]$ rank Beta $>$ clipped Gaussian $>$ reflected Gaussian, on every seed
and at both learning-rate settings. Reflection folds the draws that leave the box back inside. At $e = 0.02$ its spread is nearly the clipped Gaussian's and
its bias twice as large, and it is the slower of the two. It also lifts a quarter of the near-silent draws by more than
0.1 and keeps a 60--90\% higher world-model loss. Beta removes the bias but also cuts the variance near the bounds to a
fifth of clipping's, so the bias may not explain all of its gain (\cref{sec:supp-beta}).
 Lattice~\cite{chiappa2023lattice}, whose noise enters through
the policy's last hidden layer and so is correlated across muscles, also stays in the box when applied to the logits:
at the same scale it reaches 0.83 at the same iteration as Beta or one checkpoint later (\cref{fig:ladder}). So the bias at the bounds, rather than the noise's independence across muscles, appears to slow the
training, and Beta is the simplest noise without it.

\paragraph{Embodied exploration}
DEP-RL~\cite{schumacher2023deprl}, which drives muscles with a self-organizing controller during exploration and was
designed for overactuated systems, fails more training episodes, reaches 0.83 at the same iteration or up to two checkpoints
later (3--7\% later on average), and does not help either
as published or during the warm-up only.

\paragraph{World model and optimizer}
Ten world-model batches per iteration instead of five cost 9\% per iteration for no consistent gain in time. Muon momentum 0.9
instead of 0.95 ends lower on every seed. Removing the 200-iteration world-model-only warm-up reaches 0.83 earlier on every seed but 0.9 later on two of three, so we keep it.

\paragraph{Control rate}
Halving the control rate with the exploration noise unchanged reaches 0.83 as early but ends lower on every seed
and tracks all 108 clips less often, which rescaling the noise by \cref{eq:rate} fixes. Keeping the 100~Hz step
settings fixes it too, but at 0.28~s per iteration and 10 minutes to 0.83 (\cref{sec:supp-horizon} varies the horizon
alone).

\subsection{Tracking accuracy}
\label{sec:accuracy}

\begin{table*}[t]
  \centering
  \footnotesize
  \setlength{\tabcolsep}{2.5pt}
  \begin{tabular}{@{}llrrrrrrrrr@{}}
    \toprule
    & model, data, training & return & failed & joint MAE & joint MAE & joint RMSE & MPJPE & body RMSE & global & root \\
    & & & & (indep., \dg) & (all, \dg) & (indep., \dg) & (mm) & (cm) & (mm) & (mrad) \\
    \midrule
    baseline & \msh, KIT, 10.1 h & 0.918 & 3 & 5.14 & 2.83 & 8.82 & 22.4 & 3.12 & 136 & 24 \\
    ours, 100 Hz & \msh, KIT, 20 min & \best{0.938} & \best{1.5} & 4.65 & 2.53 & 8.17 & 18.3 & \best{2.66} & \best{98} & 14 \\
    ours, 50 Hz & \msh, KIT, 20 min & 0.936 & 3.0 & \best{3.37} & \best{1.91} & \best{5.36} & \best{18.2} & 2.68 & 123 & 14 \\
    ours, 50 Hz, train clips & \msh, KIT, 20 min & 0.954 & 5.3 / 972 & 3.14 & 1.75 & 4.96 & 16.2 & 2.33 & 107 & 9 \\
    \midrule
    baseline & \mfb, KIT, 9.6 h & 0.903 & \best{0} & 5.86 & 2.75 & 8.99 & 29.8 & 4.24 & 150 & 33 \\
    ours, 50 Hz & \mfb, KIT, 20 min & 0.906 & 1.0 & 5.45 & 2.59 & 8.44 & 28.0 & 4.15 & 149 & 30 \\
    ours, 50 Hz, refined & \mfb, KIT, 20 min & \best{0.931} & \best{0} & \best{5.02} & \best{2.37} & \best{7.96} & \best{23.6} & 3.58 & \best{88} & \best{19} \\
    MuscleMimic$^\ddagger$~\cite{li2026musclemimic} & \mfb, KIT, $\sim$3 d & 0.888 & 3.7 & 5.93 & 2.80 & 9.13 & 25.0 & \best{3.57} & 126 & \\
    \midrule
    MS-Emulator~\cite{wei2026msemulator} & \msh, one run, 7 h & & & & $\le$2$^\dagger$ & & & & & \\
    NDM Gym SuperTrack~\cite{dou2026ndmgym} & MuscleVAE char., 3 h & & & & & & 38.1$^\P$ & & & \\
    NDM Gym SuperMimic~\cite{dou2026ndmgym} & MuscleVAE char., 3 h & & & & & & 65.6$^\P$ & & & \\
    \bottomrule
  \end{tabular}
  \caption{\textbf{Tracking errors on the 108 held-out KIT clips.} Ours are means over seeds, root is the root
  orientation error. $^\ddagger$MuscleMimic's released checkpoint, evaluated by us with our terminations and metrics, its
  actions sampled as its own evaluation does (deterministic: 19 failed clips, \cref{tab:musclemimic};
  \cref{sec:supp-musclemimic}). The last block lists the numbers other whole-body muscle controllers report on other
  settings, $^\dagger$over all joints of one trajectory and $^\P$on another model and data (\cref{sec:supp-prior}).}
  \label{tab:accuracy}
\end{table*}

\Cref{tab:accuracy} measures the 20-minute policies with the error definitions of prior work, on the full test split.
At 50~Hz they fail $3.0 \pm 2.6$ of the 108 held-out clips, as many as the baseline
after 10 hours, and have a lower error in every metric. Against 100~Hz they are as precise in position and more precise
in joint angle (3.4\dg against 4.7\dg), with more global drift. Against MS-Emulator and NDM Gym the comparison is indicative, not controlled (\cref{sec:supp-prior}). We evaluate MuscleMimic's
released \mfb tracker ourselves, on the same split, terminations and metrics. Our 20-minute \mfb policies (\cref{sec:mfb}) fail fewer clips than its three days of PPO, at a higher return and lower joint-angle errors but higher position errors (MPJPE 28.0 against 25.0\mm, body RMSE 4.15 against 3.57\cm). Test-time refinement brings the MPJPE below its own (23.6 against 25.0\mm).

\subsection{Test-time refinement}
\label{sec:refine}

\begin{table}[t]
  \centering
  \footnotesize
  \setlength{\tabcolsep}{2.5pt}
  \begin{tabular}{@{}lrrrrrr@{}}
    \toprule
    & return & failed & joint MAE & MPJPE & global & root \\
    \midrule
    policy ($K{=}0$) & 0.936 & 3.0 & 3.37\dg & 18.2 & 123 & 14.4 \\
    $K{=}2$, $\eta{=}0.01$ & 0.962 & 0 & 3.12\dg & 14.5 & 63 & 4.8 \\
    $K{=}5$, $\eta{=}0.01$ & 0.965 & 0 & 2.97\dg & 13.5 & 58 & 4.0 \\
    $K{=}5$, $\eta{=}0.02$ & 0.961 & 0 & 2.94\dg & 13.5 & 57 & 4.5 \\
    $K{=}10$, $\eta{=}0.005$ & \best{0.967} & 0 & 2.95\dg & 13.5 & 59 & \best{3.7} \\
    $K{=}5$, $\eta{=}0.01$, $H{=}8$ & 0.959 & 0 & 3.20\dg & 15.7 & 82 & 4.0 \\
    $K{=}5$, $\eta{=}0.01$, last & 0.956 & 0 & 3.11\dg & 14.0 & 62 & 5.2 \\
    \bottomrule
  \end{tabular}
  \caption{\textbf{Test-time action refinement through the world model.} The 20-minute 50~Hz policies on the 108 held-out clips, mean over seeds (MPJPE and global in mm, root in mrad).}
  \label{tab:refine}
\end{table}

\Cref{tab:refine} applies \cref{eq:refine} to the 50~Hz policies of \cref{tab:accuracy}. Five gradient steps on the
imagined return lower the real MPJPE by 26\%, the global MPJPE by 53\%, the joint MAE by 12\% and the root orientation error by 72\%, with no clip
failing on any seed. Evaluated with the refinement, at about four times the cost per control step, the training reaches
0.83 after 5.1 instead of 6.3~minutes, every seed at its first checkpoint evaluated (\cref{fig:curves}). Two steps give
most of it and five saturate. The full training horizon
matters: $H = 8$ loses over a third of the global gain, so the improvement comes from planning over 0.3~s rather than from a one-step correction.
Over these 16 steps the world model's open-loop error is a third of the policy's tracking error (\cref{sec:supp-wm}).
It is therefore accurate enough to plan on, most of all against drift (\cref{fig:lafan-qual}). Unlike the policy, the
refinement sees the root's horizontal error, where its largest gains are. With this error in the policy's goal on LAFAN1
it still halves the global MPJPE (\cref{sec:supp-lafan}), so part of the remaining error comes from optimization rather
than missing information.

\paragraph{Refinement in training}
Regressing the policy onto refined actions in training, as PODS~\cite{mora2021pods} and
$\nabla$Mimic~\cite{lu2026nablamimic} do with exact gradients, slowed training down both ways we tried (\cref{sec:supp-pods}). We attribute this to imagined windows being cheap and
to a learned model's gradient leading the policy where the model is poor.

\subsection{\mfb and the baseline}
\label{sec:mfb}

On \mfb, with MuscleMimic's own retargets, the baseline reaches 0.83 after 6.0 hours and the recipe tuned on \msh, run
unchanged, after 10.5~minutes, 34$\times$ sooner (\cref{fig:curves}). Its 20-minute policies are as accurate as the
baseline after 9.6 hours (\cref{tab:accuracy}), and refined at test time they fail no clip. Trained on to one hour, the
policies' MPJPE levels off after 30--40 minutes while the refined one keeps falling, to 17.1\mm, which suggests that the world model
and critic keep improving after the policy stops.

\subsection{Limitations}
\label{sec:limits}

Our main evaluation uses KIT clips of about 5~s. On LAFAN1~\cite{harvey2020lafan1}, 77 clips of 1.5--5 minutes
retargeted to both whole-body musculoskeletal models (\cref{sec:supp-lafan}), the recipe trained for an hour fails all
seven held-out clips on \msh after 7--25~s. Every failure is the root drifting 1~m from the reference's horizontal
position while the posture is tracked. MuscleMimic's goal does not tell the policy where the reference is relative to
the body. Adding the reference root at the goal's frames, in the body's root frame, lets the policy track on average
2.5 of the seven clips to their end. With its actions refined through the world model (\cref{sec:refine}, $K = 5$) it
tracks 5.5, against 4 refined without the new goal (\cref{tab:lafan-relroot}). \msh has no collision geometry on its
arms and hands, so it cannot track hand-supported motion. \mfb has this geometry, but after an hour of training its
policy alone still fails every held-out clip, by its posture as well as by drift. Refined, it tracks 5 of the seven to
their end (\cref{tab:lafan-refine}). Under the protocol of \cref{sec:supp-emg}, our policies' activations correlate
with walking EMG as weakly as those of MuscleMimic's released tracker (about 0.1 over eleven leg muscles), and both
co-contract. An effort penalty lowers the co-contraction at a cost in tracking (\cref{tab:emg}).

\section{Conclusion}
\label{sec:conclusion}

Whole-body motion tracking on a 700-muscle human can be learned in minutes. We extended SuperMimic to two musculoskeletal
models with MuJoCo's activation dynamics inside the world model and measured, step by step, what
brings the training from 3.9 hours to 6.3 minutes on one GPU. The G1 method changes, with three settings replaced for
muscles, halve the iterations. The systems changes, among them our observation without muscle geometry and our
collection concurrent with the updates, cut the time per iteration from 1.89 to 0.22~s with a nearly unchanged learning
curve. Our Beta noise and 50~Hz control take the time from 13.7 to 6.3 minutes. The 20-minute policies track the held-out KIT clips at 18\mm and
3.4\dg. Refining the actions through the world model at test time lowers the MPJPE by a further quarter with no held-out clip
failing, and with the reference root added to the goal carries on average 5.5 of seven minutes-long clips to their end.
The learned model beats the policy that uses it, yet training on its refined actions did not help.

\section*{Acknowledgments}

The SuperMimic implementation builds on Chong Zhang's supermimic\_vibed~\cite{zhang2026supermimicvibed}; the fast
training recipe adopts Fuyi Chen's BlinkMimic~\cite{blinkmimic2026} and InstantTrack~\cite{instanttrack2026}, which it
builds on. \mfb, its tracking task and the retargeted KIT motions come from MuscleMimic~\cite{li2026musclemimic} and
MyoSuite~\cite{caggiano2022myosuite}; \msh from its authors~\cite{zuo2024mshuman700}; the simulation runs on
mjlab~\cite{zakka2026mjlab} and MuJoCo-Warp~\cite{todorov2012mujoco,mujocowarp}. The motions are from the KIT set of
AMASS~\cite{mahmood2019amass,mandery2016kit} and from LAFAN1~\cite{harvey2020lafan1} (CC BY-NC-ND~4.0; its retargets
are not redistributed).

\section*{LLM Usage Statement}

The authors declare that LLMs were used during the preparation of this manuscript to assist with language polishing
and code writing. All suggestions were reviewed and revised by the authors, who take full responsibility for the final
content.

{
    \small
    \bibliographystyle{ieeenat_fullname}
    \bibliography{main}

\begin{thebibliography}{62}
\providecommand{\natexlab}[1]{#1}
\providecommand{\url}[1]{\texttt{#1}}
\expandafter\ifx\csname urlstyle\endcsname\relax
  \providecommand{\doi}[1]{doi: #1}\else
  \providecommand{\doi}{doi: \begingroup \urlstyle{rm}\Url}\fi

\bibitem[Araujo et~al.(2025)Araujo, Ze, Xu, Wu, and Liu]{araujo2025gmr}
Joao~Pedro Araujo, Yanjie Ze, Pei Xu, Jiajun Wu, and C.~Karen Liu.
\newblock Retargeting matters: General motion retargeting for humanoid motion
  tracking.
\newblock \emph{arXiv preprint arXiv:2510.02252}, 2025.

\bibitem[Berg et~al.(2023)Berg, Caggiano, and Kumar]{berg2023sar}
Cameron Berg, Vittorio Caggiano, and Vikash Kumar.
\newblock {SAR}: Generalization of physiological agility and dexterity via
  synergistic action representation.
\newblock In \emph{Robotics: Science and Systems}, 2023.

\bibitem[Boo et~al.(2025)Boo, Seo, Kim, and Koo]{boo2025gait120}
Junyo Boo, Dongwook Seo, Minseung Kim, and Seungbum Koo.
\newblock Comprehensive human locomotion and electromyography dataset:
  {Gait120}.
\newblock \emph{Scientific Data}, 12:\penalty0 1023, 2025.

\bibitem[Caggiano et~al.(2022)Caggiano, Wang, Durandau, Sartori, and
  Kumar]{caggiano2022myosuite}
Vittorio Caggiano, Huawei Wang, Guillaume Durandau, Massimo Sartori, and Vikash
  Kumar.
\newblock {MyoSuite}: A contact-rich simulation suite for musculoskeletal motor
  control.
\newblock \emph{arXiv preprint arXiv:2205.13600}, 2022.

\bibitem[Caggiano et~al.(2023)Caggiano, Dasari, and Kumar]{caggiano2023myodex}
Vittorio Caggiano, Sudeep Dasari, and Vikash Kumar.
\newblock {MyoDex}: A generalizable prior for dexterous manipulation.
\newblock In \emph{Int. Conf. Mach. Learn.}, 2023.

\bibitem[Chen(2026)]{blinkmimic2026}
Fuyi Chen.
\newblock {BlinkMimic}: Whole-body motion tracking for the {Unitree G1} in
  minutes on one {GPU}.
\newblock \url{https://github.com/fuyichen0725/blinkmimic}, 2026.

\bibitem[Chiappa et~al.(2023)Chiappa, Marin~Vargas, Huang, and
  Mathis]{chiappa2023lattice}
Alberto~Silvio Chiappa, Alessandro Marin~Vargas, Ann~Zixiang Huang, and
  Alexander Mathis.
\newblock Latent exploration for reinforcement learning.
\newblock In \emph{Adv. Neural Inform. Process. Syst.}, 2023.

\bibitem[Choi et~al.(2026)Choi, Leem, and Won]{choi2026instantmimic}
Ikjun Choi, Geonho Leem, and Jungdam Won.
\newblock {InstantMimic}: A high performance system for learning physics-based
  skills in seconds.
\newblock In \emph{SIGGRAPH Asia Conference Papers}, 2026.
\newblock arXiv:2609.09821.

\bibitem[Chou et~al.(2017)Chou, Maturana, and Scherer]{chou2017beta}
Po-Wei Chou, Daniel Maturana, and Sebastian Scherer.
\newblock Improving stochastic policy gradients in continuous control with deep
  reinforcement learning using the {Beta} distribution.
\newblock In \emph{Int. Conf. Mach. Learn.}, pages 834--843, 2017.

\bibitem[Clavera et~al.(2020)Clavera, Fu, and Abbeel]{clavera2020maac}
Ignasi Clavera, Violet Fu, and Pieter Abbeel.
\newblock Model-augmented actor-critic: Backpropagating through paths.
\newblock In \emph{Int. Conf. Learn. Represent.}, 2020.

\bibitem[Delp et~al.(2007)Delp, Anderson, Arnold, Loan, Habib, John,
  Guendelman, and Thelen]{delp2007opensim}
Scott~L. Delp, Frank~C. Anderson, Allison~S. Arnold, Peter Loan, Ayman Habib,
  Chand~T. John, Eran Guendelman, and Darryl~G. Thelen.
\newblock {OpenSim}: Open-source software to create and analyze dynamic
  simulations of movement.
\newblock \emph{IEEE Transactions on Biomedical Engineering}, 54\penalty0
  (11):\penalty0 1940--1950, 2007.

\bibitem[Feng et~al.(2023)Feng, Xu, and Liu]{feng2023musclevae}
Yusen Feng, Xiyan Xu, and Libin Liu.
\newblock {MuscleVAE}: Model-based controllers of muscle-actuated characters.
\newblock In \emph{SIGGRAPH Asia Conference Papers}, 2023.

\bibitem[Fujita and Maeda(2018)]{fujita2018clipped}
Yasuhiro Fujita and Shin-ichi Maeda.
\newblock Clipped action policy gradient.
\newblock In \emph{Int. Conf. Mach. Learn.}, pages 1597--1606, 2018.

\bibitem[Fussell et~al.(2021)Fussell, Bergamin, and
  Holden]{fussell2021supertrack}
Levi Fussell, Kevin Bergamin, and Daniel Holden.
\newblock {SuperTrack}: Motion tracking for physically simulated characters
  using supervised learning.
\newblock \emph{ACM Trans. Graph.}, 40\penalty0 (6), 2021.

\bibitem[Georgiev et~al.(2024)Georgiev, Srinivasan, Xu, Heiden, and
  Garg]{georgiev2024ahac}
Ignat Georgiev, Krishnan Srinivasan, Jie Xu, Eric Heiden, and Animesh Garg.
\newblock Adaptive horizon actor-critic for policy learning in contact-rich
  differentiable simulation.
\newblock In \emph{Int. Conf. Mach. Learn.}, 2024.

\bibitem[{Google DeepMind and NVIDIA}(2025)]{mujocowarp}
{Google DeepMind and NVIDIA}.
\newblock {MuJoCo Warp}: A {GPU}-accelerated version of the {MuJoCo} physics
  simulator.
\newblock \url{https://github.com/google-deepmind/mujoco_warp}, 2025.

\bibitem[Hafner et~al.(2020)Hafner, Lillicrap, Ba, and
  Norouzi]{hafner2020dreamer}
Danijar Hafner, Timothy Lillicrap, Jimmy Ba, and Mohammad Norouzi.
\newblock Dream to control: Learning behaviors by latent imagination.
\newblock In \emph{Int. Conf. Learn. Represent.}, 2020.

\bibitem[Hansen et~al.(2024)Hansen, Su, and Wang]{hansen2024tdmpc2}
Nicklas Hansen, Hao Su, and Xiaolong Wang.
\newblock {TD-MPC2}: Scalable, robust world models for continuous control.
\newblock In \emph{Int. Conf. Learn. Represent.}, 2024.

\bibitem[Harvey et~al.(2020)Harvey, Yurick, Nowrouzezahrai, and
  Pal]{harvey2020lafan1}
F{\'e}lix~G. Harvey, Mike Yurick, Derek Nowrouzezahrai, and Christopher Pal.
\newblock Robust motion in-betweening.
\newblock \emph{ACM Trans. Graph.}, 39\penalty0 (4), 2020.

\bibitem[He et~al.(2024)He, Zuo, Ma, and Sui]{he2024dynsyn}
Kaibo He, Chenhui Zuo, Chengtian Ma, and Yanan Sui.
\newblock {DynSyn}: Dynamical synergistic representation for efficient learning
  and control in overactuated embodied systems.
\newblock In \emph{Int. Conf. Mach. Learn.}, 2024.

\bibitem[Hill(1938)]{hill1938heat}
Archibald~Vivian Hill.
\newblock The heat of shortening and the dynamic constants of muscle.
\newblock \emph{Proceedings of the Royal Society of London. Series B,
  Biological Sciences}, 126\penalty0 (843):\penalty0 136--195, 1938.

\bibitem[{Huy-Quang-Dao}(2026)]{instanttrack2026}
{Huy-Quang-Dao}.
\newblock {InstantTrack}: Whole-body motion tracking for the {Unitree G1} in 15
  minutes on one {GPU}.
\newblock \url{https://github.com/Huy-Quang-Dao/instantTrack}, 2026.

\bibitem[{HybridRobotics}(2026)]{supermimic2026code}
{HybridRobotics}.
\newblock {SuperMimic}: Official implementation.
\newblock \url{https://github.com/HybridRobotics/supermimic}, 2026.

\bibitem[Jordan et~al.(2024)Jordan, Jin, Boza, You, Cesista, Newhouse, and
  Bernstein]{jordan2024muon}
Keller Jordan, Yuchen Jin, Vlado Boza, Jiacheng You, Franz Cesista, Laker
  Newhouse, and Jeremy Bernstein.
\newblock Muon: An optimizer for hidden layers in neural networks.
\newblock \url{https://kellerjordan.github.io/posts/muon/}, 2024.

\bibitem[Lee et~al.(2019)Lee, Park, Lee, and Lee]{lee2019scalable}
Seunghwan Lee, Moonseok Park, Kyoungmin Lee, and Jehee Lee.
\newblock Scalable muscle-actuated human simulation and control.
\newblock \emph{ACM Trans. Graph.}, 38\penalty0 (4):\penalty0 1--13, 2019.

\bibitem[Lencioni et~al.(2019)Lencioni, Carpinella, Rabuffetti, Marzegan, and
  Ferrarin]{lencioni2019data}
Tiziana Lencioni, Ilaria Carpinella, Marco Rabuffetti, Alberto Marzegan, and
  Maurizio Ferrarin.
\newblock Human kinematic, kinetic and {EMG} data during different walking and
  stair ascending and descending tasks.
\newblock \emph{Scientific Data}, 6:\penalty0 309, 2019.

\bibitem[Li et~al.(2026)Li, Wang, Ziliotto, Simos, Kovecses, Durandau, and
  Mathis]{li2026musclemimic}
Chengkun Li, Cheryl Wang, Bianca Ziliotto, Merkourios Simos, Jozsef Kovecses,
  Guillaume Durandau, and Alexander Mathis.
\newblock Towards embodied {AI} with {MuscleMimic}: Unlocking full-body
  musculoskeletal motor learning at scale.
\newblock \emph{arXiv preprint arXiv:2603.25544}, 2026.

\bibitem[Liao et~al.(2025)Liao, Truong, Huang, Gao, Tevet, Sreenath, and
  Liu]{liao2025beyondmimic}
Qiayuan Liao, Takara~E. Truong, Xiaoyu Huang, Yuman Gao, Guy Tevet, Koushil
  Sreenath, and C.~Karen Liu.
\newblock {BeyondMimic}: From motion tracking to versatile humanoid control via
  guided diffusion.
\newblock \emph{arXiv preprint arXiv:2508.08241}, 2025.

\bibitem[Liao et~al.(2026)Liao, Shao, Long, Zakka, Horowitz, Abbeel, and
  Sreenath]{liao2026supermimic}
Qiayuan Liao, Yiyang Shao, Junfeng Long, Kevin Zakka, Roberto Horowitz, Pieter
  Abbeel, and Koushil Sreenath.
\newblock {SuperMimic}: Scaling humanoid whole-body control with one
  differentiable world model.
\newblock In \emph{Conf. Robot Learn.}, 2026.
\newblock arXiv:2610.28480.

\bibitem[Liu et~al.(2025)Liu, Su, Yao, et~al.]{liu2025muon}
Jingyuan Liu, Jianlin Su, Xingcheng Yao, et~al.
\newblock Muon is scalable for {LLM} training.
\newblock \emph{arXiv preprint arXiv:2502.16982}, 2025.

\bibitem[Liu et~al.(2026)Liu, Jiang, Han, Zhong, Song, Liu, and
  Chen]{liu2026biosyngrasp}
Siyuan Liu, Bo Jiang, Lijun Han, Shanlin Zhong, Ci Song, Huaping Liu, and
  Jiahao Chen.
\newblock {BioSynGrasp}: Bio-inspired structured reinforcement learning with
  synergetic exploration and human demonstration for dexterous grasping of
  musculoskeletal robot.
\newblock \emph{IEEE Transactions on Automation Science and Engineering},
  23:\penalty0 6102--6116, 2026.

\bibitem[Lu and Chen(2026)]{lu2026nablamimic}
Weyl Lu and Yubei Chen.
\newblock {$\nabla$Mimic}: Ultra-efficient humanoid control with differentiable
  simulation.
\newblock \emph{arXiv preprint arXiv:2610.41654}, 2026.

\bibitem[Luo et~al.(2024)Luo, Jiang, Zhang, Zhu, Yu, Dominguez~Silva, Wang,
  Rouse, Zhou, Yuk, Zhou, and Su]{luo2024exoskeleton}
Shuzhen Luo, Menghan Jiang, Sainan Zhang, Junxi Zhu, Shuangyue Yu, Israel
  Dominguez~Silva, Tian Wang, Elliott Rouse, Bolei Zhou, Hyunwoo Yuk, Xianlian
  Zhou, and Hao Su.
\newblock Experiment-free exoskeleton assistance via learning in simulation.
\newblock \emph{Nature}, 630\penalty0 (8016):\penalty0 353--359, 2024.

\bibitem[Luo et~al.(2023)Luo, Cao, Winkler, Kitani, and Xu]{luo2023phc}
Zhengyi Luo, Jinkun Cao, Alexander Winkler, Kris Kitani, and Weipeng Xu.
\newblock Perpetual humanoid control for real-time simulated avatars.
\newblock In \emph{Int. Conf. Comput. Vis.}, pages 10895--10904, 2023.

\bibitem[Luo et~al.(2026)Luo, Yuan, Wang, Li, Casta{\~n}eda, Chen, Cao, Li,
  Minor, Ben, Park, Sami, Wang, Da, Ding, Hogg, Song, Lim, Jeong, He, Xue,
  Xiao, Yuen, Kautz, Chang, Iqbal, Fan, and Zhu]{luo2026sonic}
Zhengyi Luo, Ye Yuan, Tingwu Wang, Chenran Li, Fernando Casta{\~n}eda, Sirui
  Chen, Zi-Ang Cao, Jiefeng Li, David Minor, Qingwei Ben, Jinhyung Park, David
  Sami, Zi Wang, Xingye Da, Runyu Ding, Cyrus Hogg, Lina Song, Edy Lim, Eugene
  Jeong, Tairan He, Haoru Xue, Wenli Xiao, Simon Yuen, Jan Kautz, Yan Chang,
  Umar Iqbal, Linxi Fan, and Yuke Zhu.
\newblock {SONIC}: Supersizing motion tracking for natural humanoid whole-body
  control.
\newblock \emph{Science Robotics}, 11\penalty0 (117):\penalty0 eaed4592, 2026.

\bibitem[Mahmood et~al.(2019)Mahmood, Ghorbani, Troje, Pons-Moll, and
  Black]{mahmood2019amass}
Naureen Mahmood, Nima Ghorbani, Nikolaus~F. Troje, Gerard Pons-Moll, and
  Michael~J. Black.
\newblock {AMASS}: Archive of motion capture as surface shapes.
\newblock In \emph{Int. Conf. Comput. Vis.}, 2019.

\bibitem[Mandery et~al.(2016)Mandery, Terlemez, Do, Vahrenkamp, and
  Asfour]{mandery2016kit}
Christian Mandery, {\"O}mer Terlemez, Martin Do, Nikolaus Vahrenkamp, and Tamim
  Asfour.
\newblock Unifying representations and large-scale whole-body motion databases
  for studying human motion.
\newblock \emph{IEEE Trans. Robot.}, 32\penalty0 (4), 2016.

\bibitem[Millard et~al.(2013)Millard, Uchida, Seth, and
  Delp]{millard2013flexing}
Matthew Millard, Thomas Uchida, Ajay Seth, and Scott~L. Delp.
\newblock Flexing computational muscle: Modeling and simulation of
  musculotendon dynamics.
\newblock \emph{Journal of Biomechanical Engineering}, 135\penalty0
  (2):\penalty0 021005, 2013.

\bibitem[Mittal et~al.(2025)Mittal, Roth, Tigue, et~al.]{mittal2025isaaclab}
Mayank Mittal, Pascal Roth, James Tigue, et~al.
\newblock {Isaac Lab}: A {GPU}-accelerated simulation framework for multi-modal
  robot learning.
\newblock \emph{arXiv preprint arXiv:2511.04831}, 2025.

\bibitem[{NDM Gym Contributors}(2026)]{dou2026ndmgym}
{NDM Gym Contributors}.
\newblock {NDM Gym}: {Neural Dynamics Modeling Gym}.
\newblock \url{https://github.com/Frank-ZY-Dou/ndm-gym}, 2026.

\bibitem[Peng et~al.(2018)Peng, Abbeel, Levine, and van~de
  Panne]{peng2018deepmimic}
Xue~Bin Peng, Pieter Abbeel, Sergey Levine, and Michiel van~de Panne.
\newblock {DeepMimic}: Example-guided deep reinforcement learning of
  physics-based character skills.
\newblock \emph{ACM Trans. Graph.}, 37\penalty0 (4), 2018.

\bibitem[Rajagopal et~al.(2016)Rajagopal, Dembia, DeMers, Delp, Hicks, and
  Delp]{rajagopal2016fullbody}
Apoorva Rajagopal, Christopher~L. Dembia, Matthew~S. DeMers, Denny~D. Delp,
  Jennifer~L. Hicks, and Scott~L. Delp.
\newblock Full-body musculoskeletal model for muscle-driven simulation of human
  gait.
\newblock \emph{IEEE Transactions on Biomedical Engineering}, 63\penalty0
  (10):\penalty0 2068--2079, 2016.

\bibitem[Ren et~al.(2023)Ren, Yu, Chen, Ma, Pan, and Liu]{ren2023diffmimic}
Jiawei Ren, Cunjun Yu, Siwei Chen, Xiao Ma, Liang Pan, and Ziwei Liu.
\newblock {DiffMimic}: Efficient motion mimicking with differentiable physics.
\newblock In \emph{Int. Conf. Learn. Represent.}, 2023.

\bibitem[Rudin et~al.(2022)Rudin, Hoeller, Reist, and
  Hutter]{rudin2022learning}
Nikita Rudin, David Hoeller, Philipp Reist, and Marco Hutter.
\newblock Learning to walk in minutes using massively parallel deep
  reinforcement learning.
\newblock In \emph{Conf. Robot Learn.}, pages 91--100, 2022.

\bibitem[Schulman et~al.(2017)Schulman, Wolski, Dhariwal, Radford, and
  Klimov]{schulman2017ppo}
John Schulman, Filip Wolski, Prafulla Dhariwal, Alec Radford, and Oleg Klimov.
\newblock Proximal policy optimization algorithms.
\newblock \emph{arXiv preprint arXiv:1707.06347}, 2017.

\bibitem[Schumacher et~al.(2023)Schumacher, H{\"a}ufle, B{\"u}chler, Schmitt,
  and Martius]{schumacher2023deprl}
Pierre Schumacher, Daniel H{\"a}ufle, Dieter B{\"u}chler, Syn Schmitt, and
  Georg Martius.
\newblock {DEP-RL}: Embodied exploration for reinforcement learning in
  overactuated and musculoskeletal systems.
\newblock In \emph{Int. Conf. Learn. Represent.}, 2023.

\bibitem[Simos et~al.(2025)Simos, Chiappa, and Mathis]{simos2025kinesis}
Merkourios Simos, Alberto~Silvio Chiappa, and Alexander Mathis.
\newblock {KINESIS}: Motion imitation for human musculoskeletal locomotion.
\newblock In \emph{IEEE Int. Conf. Robot. Autom.}, 2025.

\bibitem[Simos et~al.(2026)Simos, Li, Ziliotto, and Mathis]{simos2026terra}
Merkourios Simos, Chengkun Li, Bianca Ziliotto, and Alexander Mathis.
\newblock {TERRA}: Terrain-aware reconstruction, retargeting and control for
  musculoskeletal locomotion.
\newblock \emph{arXiv preprint arXiv:2609.38653}, 2026.

\bibitem[Tallec et~al.(2019)Tallec, Blier, and
  Ollivier]{tallec2019discretization}
Corentin Tallec, L{\'e}onard Blier, and Yann Ollivier.
\newblock Making deep {Q}-learning methods robust to time discretization.
\newblock In \emph{Int. Conf. Mach. Learn.}, pages 6096--6104, 2019.

\bibitem[Todorov et~al.(2012)Todorov, Erez, and Tassa]{todorov2012mujoco}
Emanuel Todorov, Tom Erez, and Yuval Tassa.
\newblock {MuJoCo}: A physics engine for model-based control.
\newblock In \emph{IEEE/RSJ Int. Conf. Intell. Robots Syst.}, 2012.

\bibitem[Wang and Ba(2020)]{wang2020poplin}
Tingwu Wang and Jimmy Ba.
\newblock Exploring model-based planning with policy networks.
\newblock In \emph{Int. Conf. Learn. Represent.}, 2020.

\bibitem[Wei et~al.(2025)Wei, Zhuang, Zhuang, and Sui]{wei2025mpc2}
Yunyue Wei, Shanning Zhuang, Vincent Zhuang, and Yanan Sui.
\newblock Motion control of high-dimensional musculoskeletal systems with
  hierarchical model-based planning.
\newblock In \emph{Int. Conf. Learn. Represent.}, 2025.

\bibitem[Wei et~al.(2026{\natexlab{a}})Wei, Zuo, and Sui]{wei2026qflex}
Yunyue Wei, Chenhui Zuo, and Yanan Sui.
\newblock Scalable exploration for high-dimensional continuous control via
  value-guided flow.
\newblock In \emph{Int. Conf. Learn. Represent.}, 2026{\natexlab{a}}.
\newblock arXiv:2601.19707.

\bibitem[Wei et~al.(2026{\natexlab{b}})Wei, Zuo, Zhuang, Gong, Liu, and
  Sui]{wei2026msemulator}
Yunyue Wei, Chenhui Zuo, Shanning Zhuang, Haixin Gong, Yaming Liu, and Yanan
  Sui.
\newblock Scaling whole-body human musculoskeletal behavior emulation for
  specificity and diversity.
\newblock \emph{arXiv preprint arXiv:2603.29332}, 2026{\natexlab{b}}.

\bibitem[Weng et~al.(2025)Weng, Li, Sobanbabu, Wang, Luo, He, Ramanan, and
  Shi]{weng2025hdmi}
Haoyang Weng, Yitang Li, Nikhil Sobanbabu, Zihan Wang, Zhengyi Luo, Tairan He,
  Deva Ramanan, and Guanya Shi.
\newblock {HDMI}: Learning interactive humanoid whole-body control from human
  videos.
\newblock \emph{arXiv preprint arXiv:2509.16757}, 2025.

\bibitem[Xu et~al.(2022)Xu, Makoviychuk, Narang, Ramos, Matusik, Garg, and
  Macklin]{xu2022shac}
Jie Xu, Viktor Makoviychuk, Yashraj Narang, Fabio Ramos, Wojciech Matusik,
  Animesh Garg, and Miles Macklin.
\newblock Accelerated policy learning with parallel differentiable simulation.
\newblock In \emph{Int. Conf. Learn. Represent.}, 2022.

\bibitem[Yao et~al.(2022)Yao, Song, Chen, and Liu]{yao2022controlvae}
Heyuan Yao, Zhenhua Song, Baoquan Chen, and Libin Liu.
\newblock {ControlVAE}: Model-based learning of generative controllers for
  physics-based characters.
\newblock \emph{ACM Trans. Graph.}, 41\penalty0 (6):\penalty0 183, 2022.

\bibitem[Zakka et~al.(2026)Zakka, Liao, Yi, Le~Lay, Sreenath, and
  Abbeel]{zakka2026mjlab}
Kevin Zakka, Qiayuan Liao, Brent Yi, Louis Le~Lay, Koushil Sreenath, and Pieter
  Abbeel.
\newblock mjlab: A lightweight framework for {GPU}-accelerated robot learning.
\newblock \emph{arXiv preprint arXiv:2601.22074}, 2026.

\bibitem[Zamora~Mora et~al.(2021)Zamora~Mora, Peychev, Ha, Vechev, and
  Coros]{mora2021pods}
Miguel~Angel Zamora~Mora, Momchil Peychev, Sehoon Ha, Martin Vechev, and
  Stelian Coros.
\newblock {PODS}: Policy optimization via differentiable simulation.
\newblock In \emph{Int. Conf. Mach. Learn.}, pages 7805--7817, 2021.

\bibitem[Zeni et~al.(2008)Zeni, Richards, and Higginson]{zeni2008gait}
J.~A. Zeni, J.~G. Richards, and J.~S. Higginson.
\newblock Two simple methods for determining gait events during treadmill and
  overground walking using kinematic data.
\newblock \emph{Gait \& Posture}, 27\penalty0 (4):\penalty0 710--714, 2008.

\bibitem[Zhang(2026)]{zhang2026supermimicvibed}
Chong Zhang.
\newblock supermimic\_vibed: An unofficial reimplementation of {SuperMimic}.
\newblock \url{https://github.com/zita-ch/supermimic_vibed}, 2026.

\bibitem[Zuo et~al.(2024)Zuo, He, Shao, and Sui]{zuo2024mshuman700}
Chenhui Zuo, Kaibo He, Jing Shao, and Yanan Sui.
\newblock Self model for embodied intelligence: Modeling full-body human
  musculoskeletal system and locomotion control with hierarchical
  low-dimensional representation.
\newblock In \emph{IEEE Int. Conf. Robot. Autom.}, 2024.

\end{thebibliography}
}
\clearpage
\setcounter{page}{1}
\maketitlesupplementary

\setcounter{section}{0}
\renewcommand{\thesection}{S.\arabic{section}}
\renewcommand{\theHsection}{S.\arabic{section}}
\setcounter{figure}{0}
\renewcommand{\thefigure}{S\arabic{figure}}
\renewcommand{\theHfigure}{S\arabic{figure}}
\setcounter{table}{0}
\renewcommand{\thetable}{S\arabic{table}}
\renewcommand{\theHtable}{S\arabic{table}}
\renewcommand{\thealgorithm}{S\arabic{algorithm}}
\renewcommand{\theHalgorithm}{S\arabic{algorithm}}

\section{Training algorithm}
\label{sec:supp-algo}

\Cref{alg:train} gives one training run of the fast recipe in the notation of \cref{sec:learning}, with the settings of
\cref{tab:hparams}. The baseline is the same loop with the collection before the updates, Gaussian noise clipped to
$[0,1]$, no policy average and no clamp of the bootstrapped values.

\begin{algorithm*}[t]
  \caption{\textbf{Training with the fast recipe} (\cref{sec:learning,sec:recipe}), shaded by CUDA stream as in
  \cref{fig:overview}.}
  \label{alg:train}
  \small
  \begin{algorithmic}[1]
    \Require $N$ environments, replay buffer $\mathcal B$, world model $f_\phi$, critic $V_\psi$, policy $\pi_\theta$, which collects, and
      its Polyak average $\bar\pi$ (rate $\tau$), which is deployed. Horizon $H$, window $k$, $B_{\mathrm{wm}}$ world-model batches
      per iteration, $I_0$ world-model-only iterations
    \State \algmark{algL}fill $\mathcal B$ with transitions of $\pi_\theta$ with exploration noise
    \For{\algmark{alg2}iteration $i = 1, 2, \dots$}
      \For{\algmark{alg3}the iteration's simulator steps, all $N$ environments at once} \Comment{collection, stream 2}
        \State $\bar e \gets \pi_\theta(o(x), g)$, \quad $e \sim \mathrm{Beta}\big(\kappa \bar e, \kappa(1 - \bar e)\big)$
          \Comment{\cref{sec:recipe} (c)}
        \State $x' \gets$ the $n$ MuJoCo-Warp substeps from $x$ under $e$, \ $d \gets$ the terminations of $x'$
        \State store $(x, e, x', d)$ in $\mathcal B$ and reset the terminated environments to reference frames, part of
          them just before recent failures
      \EndFor\algmark{alg7}
      \For{\algmark{alg8}$B_{\mathrm{wm}}$ batches} \Comment{world model, stream 1}
        \State sample $k$-step windows $(x_0, e_{0:k-1}, x_{1:k})$ of $\mathcal B$, half of them mirrored left to right
        \State $\hat x_0 \gets x_0$, \ $\hat x_{j+1} \gets f_\phi(\hat x_j, e_j)$ for $j < k$, \ then a step of $\phi$ on
          $L(\phi)$ \Comment{\cref{eq:wm,eq:wmloss}}
      \EndFor\algmark{alg11}
      \If{\algmark{alg12}$i > I_0$} \Comment{policy and critic, after the world model's warm-up}
        \State sample start states $\hat x_0$ of $\mathcal B$
        \For{$t = 0, \dots, H - 1$}
          \State $e_t \gets \pi_\theta(o(\hat x_t), g_t)$, \ $\hat x_{t+1} \gets f_\phi(\hat x_t, e_t)$, \
            $\hat r_{t+1} \gets r(\hat x_{t+1}, x^{\mathrm{ref}}_{t+1})$, \ $\hat d_{t+1}$ the softened terminations
        \EndFor
        \State a step of $\theta$ up $\nabla_\theta \hat J(\theta)$, backpropagated through the $H$ imagined steps
          \Comment{\cref{eq:imag}}
        \State $\bar\pi \gets \tau \bar\pi + (1 - \tau)\, \pi_\theta$
        \State $\hat R_t$ from the rollout, held fixed, then a few epochs of minibatch steps of $\psi$ on $L_V(\psi)$
          \Comment{\cref{eq:critic}}
      \EndIf\algmark{alg20}
    \EndFor\algmark{alg21}\algshade{streamcollect}{alg2}{alg3}{alg7}{alg8}%
      \algshade{streamwm}{alg7}{alg8}{alg11}{alg12}\algshade{streampolicy}{alg11}{alg12}{alg20}{alg21}
  \end{algorithmic}
\end{algorithm*}

\section{Hyperparameters}
\label{sec:supp-hparams}

\begin{table}[h]
  \centering
  \footnotesize
  \setlength{\tabcolsep}{2.5pt}
  \begin{tabular}{@{}lll@{}}
    \toprule
    & baseline & recipe (50 Hz) \\
    \midrule
    control rate / substeps & 100 Hz / 5 & 50 Hz / 10 \\
    environments & 4096 & 1024 \\
    sim. steps per iteration & 2.5 & 1.25 \\
    buffer (steps) & 1000 & 500 \\
    world model & 3 $\times$ 2048 & 3 $\times$ 1024 \\
    world-model batches / window & 10 / 8 & 5 / 4 \\
    world-model-only iterations & 500 & 200 \\
    imagination horizon $H$ & 32 & 16 \\
    $\gamma$ / $\lambda$ & 0.99 / 0.95 & 0.98 / 0.90 \\
    imagination batch & 4096 & 2048 \\
    learning rate & 2e-4 (Adam) & 4e-4 (Muon + Adam) \\
    policy EMA & -- & 0.995 \\
    value clamp & -- & $r_{\max}/(1-\gamma)$ \\
    exploration noise & $\mathcal N(0, 0.1^2)$ clipped & Beta, $\sigma{=}0.0707$ \\
    resets before failures & -- & 30\% \\
    mirror augmentation & -- & half of each batch \\
    muscle geometry observed & yes & no \\
    CUDA graphs & -- & updates, collection \\
    rematerialization & yes & no \\
    \bottomrule
  \end{tabular}
  \caption{\textbf{Hyperparameters.} The baseline, SuperMimic's published hyperparameters as run on \msh, and the fast recipe.}
  \label{tab:hparams}
\end{table}

\begin{figure*}[t]
  \centering
  \includegraphics[width=\linewidth]{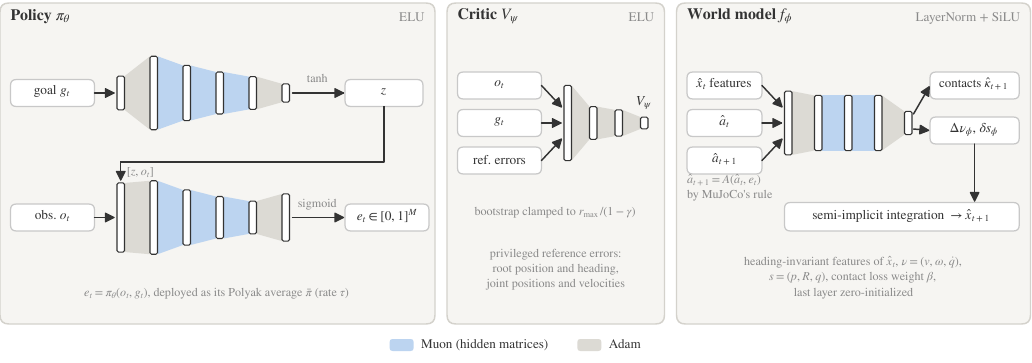}
  \caption{\textbf{The networks} (\cref{sec:learning}). Each bar is a layer, taller the wider it is, and each fill
  between two bars a weight matrix, coloured by its optimizer.}
  \label{fig:arch}
\end{figure*}

The policy is an encoder-decoder (\cref{fig:arch}): an MLP (2048, 1024, 512, 256) encodes the goal to $z \in \R^{32}$ (tanh), and
an MLP (2048, 1024, 512, 256) maps the observation and $z$ to the excitations, with its output bias set so that they start near 0.1. The critic is an MLP
(256, 128). The world-model loss (\cref{eq:wmloss}) has per-component error floors and a contact term of weight
$\beta = 1$ on the contact forces, normalized by their running mean and standard deviation. A reset starts in a 1~s bin
of a clip, chosen uniformly, weighted by return, or just before a recent failure. The updates are captured as CUDA
graphs and replayed from then on.

\section{Task, data and simulation}
\label{sec:supp-task}

\paragraph{Whole-body musculoskeletal models and simulation}
The coupled joints of \msh are the knee translations, the patella, the shoulder girdle and the wrist. Upstream it has no
collision geometry on the arms, hands or thorax, which matters for the motions it can track (\cref{sec:limits}). \mfb as
MuscleMimic configures it has its 40 finger joints and 62 finger muscles removed. Both run with MuscleMimic's solver
settings: Euler integration with 2~ms substeps, 100 solver and 50 line-search iterations.

\paragraph{MDP}
The observation is the root height and orientation, every joint position and velocity, the root velocities, every
muscle's excitation and activation (and its length, velocity and force in the baseline), four foot-contact forces and
the root-relative positions, rotations and velocities of 17 tracked sites on the trunk, head and limbs. The goal is the
reference's root-relative site positions at the current frame and four frames 0.2~s apart, the root position and
velocity deltas, and the clip phase. The reward of \cref{eq:reward} is MuscleMimic's MimicReward with its weights:
kernels on the mean squared errors of joint position and velocity, root position and velocity, and site position,
orientation and velocity, with no penalty term. The site deviation of the termination
is the mean root-relative distance. An episode also times out at the clip's end, with a bootstrapped value. There is no
domain randomization and no observation noise.

\paragraph{Data}
The KIT clips of the KINESIS split are mostly walking, turning, running and stepping, at 100~Hz. For \msh we transfer
them kinematically from \mfb, whose skeleton shares its OpenSim origins. The leg, shoulder, elbow and forearm angles
turn the same joints about the same axes and carry over unchanged. The spine and wrist joints of \msh are fitted to the
thorax's rotation relative to the pelvis and the hand's relative to the radius, and the root is lowered per clip so that
the feet touch the ground as in the source.

\section{Differentiable muscle geometry}
\label{sec:supp-replica}

The torch replica covers MuJoCo's spatial tendons as both models use them: site-to-site segments, and sphere and
cylinder wraps with side sites and inside wraps. It also covers the Hill curves of \cref{eq:hill}: active force-length
and force-velocity, and the passive force from the actuator's bias parameters. It matches MuJoCo's actuator length,
velocity and force to $10^{-9}$, $10^{-8}$ and $10^{-6}$, and the coupled joints' forward kinematics equals
\texttt{mj\_kinematics} to $10^{-9}$. Dropping the muscle geometry from the observation (\cref{sec:recipe}, (a)) halves
the policy update's time on \mfb, with the same learning curve.

\section{Time per iteration}
\label{sec:supp-time}

\Cref{tab:ladder} gives the steps of \cref{fig:ladder} in numbers, with the iteration at which each run first reaches
0.83 and the time per iteration. On \msh without muscle observations at 4096 environments (the third step), one
iteration takes 1.27~s eagerly and 0.65~s with the updates as CUDA graphs: the policy update, 60\% of the eager
iteration, is launch-bound and shrinks $5\times$ when replayed. The collection is then the bottleneck, which the graphed
collection step and the overlap with fewer environments remove. Control at 50~Hz then halves the policy steps per
simulated second, in the collection and in imagination, with the same physics substeps per iteration. Keeping the
imagined activations for the backward pass instead of recomputing them takes 12\% less time per iteration.

\begin{table}[h]
  \centering
  \footnotesize
  \setlength{\tabcolsep}{2.5pt}
  \begin{tabular}{@{}lrrr@{}}
    \toprule
    & to 0.83 & iter. & s / it. \\
    \midrule
    baseline, 4096 envs & 3.9 $\pm$ 1.0 h & 6000--7000 & 1.7--2.8 \\
    + G1 method changes & 1.8 $\pm$ 0.1 h & 3250--3750 & 1.89 \\
    + no muscle obs., graphed updates & 37.2 $\pm$ 3.5 min & 2750--3250 & 0.65 \\
    + graphed collection step & 24.2 $\pm$ 0.1 min & 3000 & 0.48 \\
    + 1024 envs, overlapped collection & 14.4 $\pm$ 0.6 min & 3250--3750 & 0.24 \\
    + no rematerialization & 13.7 $\pm$ 1.9 min & 3250--4250 & 0.22 \\
    + Beta noise$^\dagger$ & 8.7 $\pm$ 0.4 min & 2250--2500 & 0.21 \\
    + 50 Hz, settings rescaled & \best{6.3 $\pm$ 0.1 min} & 2500 & \best{0.14} \\
    \bottomrule
  \end{tabular}
  \caption{\textbf{The recipe step by step} (the coloured rows of \cref{fig:ladder}). Time and iteration of the first
  held-out return $\geq 0.83$ (mean $\pm$ standard deviation and range over three seeds), and seconds per iteration.
  $^\dagger$Simulator-side precompute compiled (6\% less time per iteration).}
  \label{tab:ladder}
\end{table}

\paragraph{Compilation and caches}
\phantomsection\label{sec:supp-compile}
Compilation and the earlier CUDA-graph captures delay the first policy-update capture to 10.6 minutes of the training
clock on a cold machine and to 1.1 minutes on a warm one. We pack a warm machine's caches and load them before
training. The times of
\cref{tab:ladder} rows 5--8 and \cref{tab:ablation} are with warm caches.

\section{Metric definitions}
\label{sec:supp-metrics}

For every control step $t$ before a clip's end or failure, with the simulated state $x_t$ and the reference frame
$x^{\mathrm{ref}}_t$: the \emph{joint MAE} is the mean of $|q_j - q^{\mathrm{ref}}_j|$ over the independent joints (37 on \msh, 33 with locked hands, 31 on \mfb,
with the wrist columns of the reference ignored). The \emph{all-joint MAE} is the same over every non-root joint (79, 71 with locked hands, 82 on \mfb), coupled ones included, with slide joints in metres (MS-Emulator's $E_{\mathrm{joint}}$). The \emph{joint RMSE} is the root
mean square of the same differences over the independent joints. With $p_i$ the world position of tracked site $i$ and
$p_0$ the pelvis, the \emph{MPJPE} is the mean over the 16 non-pelvis sites of $\|(p_i - p_0) - (p^{\mathrm{ref}}_i - p^{\mathrm{ref}}_0)\|$,
the \emph{body RMSE} the root mean
square of the same distances, and the \emph{global MPJPE} the mean of $\|p_i - p^{\mathrm{ref}}_i\|$ over all 17 sites. The
\emph{root orientation error} is the rotation angle of $(R^{\mathrm{ref}}_0)^\top R_0$. Each is pooled over all counted steps of all
clips. Averaging per clip first and then over clips (NDM Gym's convention) changes the values by under 0.1\dg, 0.3\mm
and 7\mm (global). With the pelvis counted (at zero error), the MPJPE is NDM Gym's root-relative MPJPE.

\section{Prior work in \cref{tab:accuracy}}
\label{sec:supp-prior}

MS-Emulator~\cite{wei2026msemulator} emulates one running trajectory after 7 hours of PPO on one RTX~5090 at a
joint-angle error of at most 2\dg, the MAE over all 79 joints of \msh as in our all-joint column. It reports at most
2\dg on walking and at most 7\dg and about 6\cm on highly dynamic motions. The two NDM Gym rows~\cite{dou2026ndmgym}
are the root-frame MPJPE of SuperTrack and SuperMimic on MuscleVAE's character, a different whole-body musculoskeletal
model and data, at the best checkpoint within 3~h over three seeds. Neither is evaluated on our clips, so both
comparisons are indicative only.

\section{MuscleMimic's released checkpoint}
\label{sec:supp-musclemimic}

The MuscleMimic row of \cref{tab:accuracy} is the \mfb checkpoint MuscleMimic releases, \texttt{mm-fullbody-base}. Its
policy, a residual MLP of about 50M parameters, was trained with PPO on the 972 training clips of the same split for 3.2
billion environment steps in 8192 environments (about three days on an H100 at its reported throughput). We evaluate it
with its own code and MJX-Warp simulator on its GMR retargets, every held-out clip once from its first frame, with its
training termination, which is also ours (site deviation 1~m, root 1~m, root orientation 1~rad). We add our error
definitions (\cref{sec:supp-metrics}) to its per-step diagnostics and recompute the normalized return from its reward
components with our weights. Two details decide whether the checkpoint runs at all. First, its evaluation tool sizes
the contact buffer for the number of environments it creates, not for the number it runs. With the two equal, its MJX
and CPU MuJoCo evaluations agree; otherwise every clip falls. Second, the checkpoint runs only on the model version it was trained on
(\texttt{musclemimic-models} 1.0.4). Version 1.0.6, which we train on, flips the sign of the left knee's coupling
polynomials, and on it every clip fails. Rebuilt from the same joint angles, the two versions' references are 1.2\mm
apart in MPJPE, too little to explain the MPJPE gap of \cref{tab:accuracy}.

\Cref{tab:musclemimic} compares three protocols. MuscleMimic's evaluation samples the actions by default, and its
policy is much less accurate with deterministic actions: 19 failures instead of 3.7, and an MPJPE of 43.0 instead of
25.0\mm. With its own validation termination and error definitions (RMS over all joints, root-relative site error per
coordinate), its joint and site errors are 5.86\dg and 1.87\cm, and it fails 2 clips. Its paper's table reports
6.63\dg, 2.28\cm and 92.6\% success, from start frames and a checkpoint we do not know.

\begin{table}[t]
  \centering
  \footnotesize
  \setlength{\tabcolsep}{3pt}
  \begin{tabular}{@{}lrrrrrr@{}}
    \toprule
    actions, termination & return & failed & MAE & RMSE & MPJPE & global \\
    \midrule
    sampled, ours & 0.888 & 3.7 & 5.93 & 9.13 & 25.0 & 126 \\
    deterministic, ours & 0.785 & 19 & 7.61 & 11.27 & 43.0 & 200 \\
    sampled, 0.5~m site & 0.893 & 2 & 6.00 & 9.27 & 26.0 & 141 \\
    \bottomrule
  \end{tabular}
  \caption{\textbf{MuscleMimic's released \mfb checkpoint} on the 108 held-out clips. Columns as in \cref{tab:accuracy}.
  First row: mean over evaluation seeds (return $\pm$0.004, failures $3.7 \pm 0.6$).}
  \label{tab:musclemimic}
\end{table}

\section{The first runs of the G1 recipe on \msh}
\label{sec:supp-firstruns}

With BlinkMimic's whole recipe at 4096 environments, the three settings of \cref{sec:recipe} failed on \msh. (i) The
critic diverged from the first policy updates while the world model learned normally: its imagined return reached 170,
where 12.1 is attainable. The bootstrapped-value clamp and a 200-iteration warm-up removed the divergence. (ii) The run
then led the baseline until iteration 1000 and slid from there to a held-out return of 0.15--0.20 with every clip
failing, as the share of imagined trajectories at the state-gradient clip rose from 0 to 13\%. A run without the
evaluation-guided resets slid the same way. Dropping the $4\times$ policy learning-rate phase removed the slide: the
run reached 0.83 after 1.8 hours, against 4.3 hours for the baseline of \cref{fig:curves}. (iii) Every probe of the
evaluation-guided resets failed within 2--3~s of its start, and the probe sweeps took about half of the training time.

\section{Ablations}
\label{sec:supp-ablation}

\Cref{tab:ablation} gives the runs behind the grey rows of \cref{fig:ladder} in the figure's order, with the steps they
branch from as references, and Lattice repeated on the 50~Hz recipe.

\begin{table}[t]
  \centering
  \footnotesize
  \setlength{\tabcolsep}{3pt}
  \begin{tabular}{@{}lccl@{}}
    \toprule
    change & min to 0.83 & best & \\
    \midrule
    \multicolumn{4}{@{}l}{\emph{100 Hz, 1024 envs, overlapped collection, clipped Gaussian}} \\
    reference & 13.7 $\pm$ 1.9 & 0.918 $\pm$ 0.006 & \\
    base learning rate 6e-4 & stall & 0.728 $\pm$ 0.310 & $\times$ \\
    2$\times$ policy LR to iter. 800 & \multicolumn{2}{c}{half of the runs stall} & $\times$ \\
    no world-model warm-up & 12.0 $\pm$ 0.5 & 0.914 $\pm$ 0.008 & $\circ$ \\
    10 world-model batches & 12.5 $\pm$ 0.6 & 0.919 $\pm$ 0.004 & $\circ$ \\
    Muon momentum 0.9 & 13.7 $\pm$ 0.5 & 0.906 $\pm$ 0.006 & $\times$ \\
    reflected Gaussian noise & never & 0.759 $\pm$ 0.052 & $\times$ \\
    \textbf{Beta noise} & \best{8.5 $\pm$ 0.0} & \best{0.943 $\pm$ 0.001} & \checkmark \\
    \midrule
    \multicolumn{4}{@{}l}{\emph{+ Beta noise}$^\dagger$} \\
    reference & 8.7 $\pm$ 0.4 & 0.938 $\pm$ 0.002 & \\
    Lattice noise & 8.9 $\pm$ 0.2 & 0.941 $\pm$ 0.002 & $\circ$ \\
    DEP-RL, as published & 9.3 $\pm$ 0.3 & 0.937 $\pm$ 0.000 & $\circ$ \\
    DEP-RL in the warm-up only & 9.0 $\pm$ 0.2 & 0.936 $\pm$ 0.001 & $\circ$ \\
    50 Hz, noise 0.1 & 6.1 $\pm$ 0.6 & 0.939 $\pm$ 0.000 & $\circ$ \\
    50 Hz, 100 Hz step settings & 9.8 $\pm$ 0.2 & 0.942 $\pm$ 0.003 & $\circ$ \\
    \textbf{50 Hz, noise 0.0707} & \best{6.3 $\pm$ 0.1} & \best{0.945 $\pm$ 0.000} & \checkmark \\
    \midrule
    \multicolumn{4}{@{}l}{\emph{+ 50 Hz, settings rescaled (the recipe)}} \\
    reference & 6.3 $\pm$ 0.1 & 0.945 $\pm$ 0.000 & \\
    Lattice noise & 6.4 $\pm$ 0.3 & 0.946 $\pm$ 0.001 & $\circ$ \\
    \bottomrule
  \end{tabular}
  \caption{\textbf{The ablations of \cref{fig:ladder}.} Minutes to the first held-out return of 0.83 and best held-out return
  within 25 minutes (20 in the lower two blocks), mean $\pm$ standard deviation over seeds. \checkmark adopted, $\times$
  dropped, $\circ$ no gain. \emph{Stall}: a seed's held-out return stays at 0.3--0.4. \emph{Never}: no seed
  reaches 0.83 within 25 minutes. $^\dagger$Precompute compiled (\cref{tab:ladder}).}
  \label{tab:ablation}
\end{table}

\section{Beta exploration noise}
\label{sec:supp-beta}

On a trained \msh policy, 26\% of the excitations are below 0.05. Of the clipped Gaussian's draws, 14.9\% fall below
0, so for the near-silent quarter the executed excitation averages 0.050 instead of the intended 0.0185. Any noise that stays in $[0,1]$ with mean $e$ has a
standard deviation of at most $\sqrt{e(1-e)}$; Beta's is $\sigma$ at $e = 1/2$, as set in \cref{sec:recipe}. The table
below compares the noises at $e = 0.02$; its last column is the share of draws that lift a near-silent muscle by more than 0.1.

\begin{center}
  \footnotesize
  \setlength{\tabcolsep}{4pt}
  \begin{tabular}{@{}lccc@{}}
    \toprule
    noise & bias & std.\ dev. & lift $> 0.1$ \\
    \midrule
    clipped Gaussian & 0.031 & 0.065 & 16\% \\
    reflected Gaussian & 0.061 & 0.062 & 24\% \\
    Beta & 0 & 0.028 & 1.4\% \\
    \midrule
    truncated Gaussian, scale 0.1 & 0 & 0.019 & \\
    clipped Gaussian, $\sigma = 0.037$ & 0.007 & 0.028 & \\
    \bottomrule
  \end{tabular}
\end{center}

Reflection keeps nearly the spread of clipping and doubles its bias. Neither of the obvious controls, the last two
rows, separates the bias from the spread near the bounds: the truncated Gaussian, shifted to mean $e$, has no bias but
less spread than Beta, and the clipped Gaussian with Beta's spread at $e = 0.02$ keeps some bias and has only a third of
Beta's spread at $e = 1/2$.

Lattice~\cite{chiappa2023lattice} adds to a stochastic policy's actions an independent noise term and one correlated
across actions through the output layer, both proportional to the last hidden features and of learned scale. Our
policy is deterministic and its noise only explores, so there is no scale to learn: we apply both terms to the
excitation logits at every step, with equal variance, set so that the excitation's standard deviation at $e = 1/2$ is
Beta's $\sigma$. Like Beta, this noise never leaves $[0,1]$, but near the bounds its spread shrinks as
$e(1-e)$ rather than $\sqrt{e(1-e)}$. At 100~Hz it trails Beta's held-out return by 0.01--0.06 at iterations 2250 and
2500.

\section{Imagination horizon}
\label{sec:supp-horizon}

The policy learns through $H$ imagined steps with a critic bootstrap at the end, as in SHAC~\cite{xu2022shac}. The
recipe's $H = 16$ simply keeps the 0.32~s horizon of the 100~Hz baseline at 50~Hz. AHAC~\cite{georgiev2024ahac} adapts
$H$ during training, which accounts for most of its gain over SHAC on a simulated quadruped. \Cref{tab:horizon} varies
only $H$, over seeds, with $\gamma$, $\lambda$ and the rest of the recipe unchanged. With 8 steps an iteration is
cheaper, but training is fragile: one seed stalls, its critic loss failing to fall, and the other seeds trail $H = 16$
at every checkpoint. With 24 steps the policy learns as much per iteration, its held-out return within 0.01 of
$H = 16$'s from iteration 2500 on, but each iteration takes longer, so it reaches 0.83 and 0.90 later, with the same
return and error at 10 minutes. $H = 16$ thus sits at the knee, and adapting $H$ has little to gain here.
None of the three horizons shows the exploding gradients that a short horizon is meant to avoid: at most 0.25\% of the
imagined trajectories reach the state-gradient clip, against 13\% in the slide of \cref{sec:supp-firstruns}.

\begin{table}[h]
  \centering
  \footnotesize
  \setlength{\tabcolsep}{2.5pt}
  \begin{tabular}{@{}rccccc@{}}
    \toprule
    $H$ & s/iter. & to 0.83 & to 0.90 & return & MPJPE (mm) \\
    \midrule
    8 & 0.130 & stall & never & 0.611 $\pm$ 0.397 & 59.0 $\pm$ 42.0 \\
    \textbf{16} & 0.143 & \best{6.1 $\pm$ 0.0} & \best{7.9 $\pm$ 0.0} & \best{0.923 $\pm$ 0.004} & 21.0 $\pm$ 0.8 \\
    24 & 0.165 & 6.9 $\pm$ 0.0 & 8.3 $\pm$ 0.0 & 0.921 $\pm$ 0.004 & \best{20.2 $\pm$ 0.3} \\
    \bottomrule
  \end{tabular}
  \caption{\textbf{The imagination horizon alone} (the recipe: 16 steps of 20~ms). Mean $\pm$ standard deviation over
  seeds of the seconds per iteration, the minutes to held-out returns of 0.83 and 0.90, and
  the return and MPJPE at 10 minutes. \emph{Stall}: a seed's held-out return stays at 0.3--0.4. \emph{Never}: no seed
  reaches 0.90 within 10 minutes.}
  \label{tab:horizon}
\end{table}

\section{The collection as a CUDA graph}
\label{sec:supp-collector}

During collection, a thin wrapper replaces the environment's step and reset and does the same work on the same tensors
in the same order. Beyond the parts listed in \cref{sec:recipe}~(b), the graph also holds MuJoCo-Warp's solver loop,
the buffer writes, the normalizer updates and the episode resets. Warp's capture nests in torch's, so the whole step is
one graph and needs no host synchronization. With the overlap, the collection keeps its own
copies of the policy, the normalizers and the random state. Its 1024 environments collect $4\times$ less fresh data per
iteration than 4096 and reach 0.83 after 3250--3750 iterations instead of 3000 (\cref{tab:ladder}). Tests compare the
graphed and the eager step over 40 steps with failures and time-outs: they agree
to $10^{-5}$ on the CPU and within MuJoCo-Warp's run-to-run noise on the GPU.

\section{Mirror maps}
\label{sec:supp-mirror}

Skeletal segments, muscles and contact values pair by name. Each joint's partner and sign come from the forward
kinematics: a joint moved by a small angle and then mirrored must match its partner moved by plus or minus that angle.
When training starts, every map is checked on random poses through the kinematics and the muscle geometry. The
residuals measure the models' own asymmetry. \msh's skeleton is symmetric to float precision and \mfb's to 2\mm. Muscle
paths are symmetric to well under 1\mm, except a few segments and one trunk muscle, whose
upstream path is 3--4\cm asymmetric on both models.

\section{World-model accuracy}
\label{sec:supp-wm}

Each of the three 20-minute 50~Hz \msh policies (\cref{tab:accuracy}) tracks the 108 held-out clips once in the
simulator. Every recorded state starts a world-model rollout of up to 32 steps, whose state after $k$ steps is compared with the simulator's. In open loop the world model replays the executed excitations. In closed loop the policy
acts on the imagined states, as in the policy update.

\begin{figure*}[t]
  \centering
  \includegraphics[width=\linewidth]{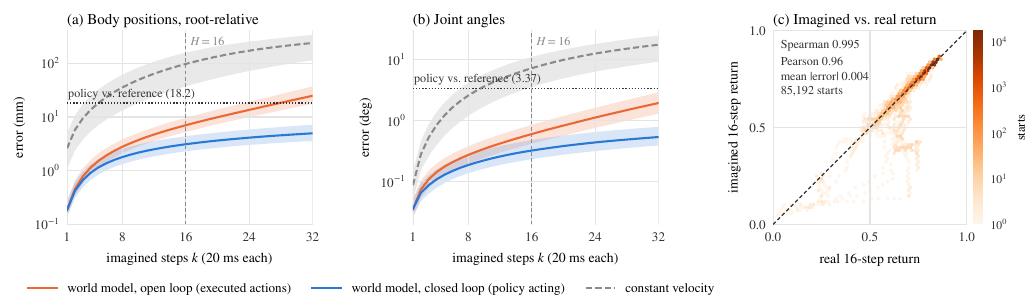}
  \caption{\textbf{World-model accuracy on the held-out clips} (three seeds, every recorded state a start). (a,
  b)~Error against the simulator $k$ steps ahead: median (mean over seeds) and interquartile range. Dotted: the
  policies' tracking error. (c)~Imagined against simulated 16-step return from the same states, normalized.}
  \label{fig:wm-accuracy}
\end{figure*}

Over the 16 steps of the training horizon (\cref{fig:wm-accuracy}a, b), the median open-loop error reaches 7.0\mm in
root-relative body position and 0.61\dg in joint angle. That is a third of the policies' tracking error (18.2\mm,
3.37\dg) and a fourteenth of a constant-velocity extrapolation's. Beyond the horizon it grows faster. In
closed loop it is 3.1\mm at 16 steps. From the same states, the imagined and the simulated
16-step returns have a Spearman rank correlation of 0.995 (\cref{fig:wm-accuracy}c) and a mean absolute difference of
0.004, 0.5\% of the mean. Near failures the model is pessimistic: where the real return is below 0.5, the imagined one is 0.04 lower on
average.

\section{Refinement over seeds}
\label{sec:supp-refine}

\Cref{tab:refine-seeds} repeats \cref{tab:refine} with the standard deviations over seeds and gives the refinement's
cost per control step. $K$, $\eta$ and $H$ were chosen on these held-out clips, but the three settings with $K \geq 5$
and $H = 16$ differ by 0.006 in return and not at all in MPJPE.

\begin{table}[h]
  \centering
  \footnotesize
  \setlength{\tabcolsep}{1pt}
  \begin{tabular}{@{}lrrrrr@{}}
    \toprule
    & return & failed & MAE (\dg) & MPJPE & global \\
    \midrule
    $K{=}0$ & 0.936$\pm$0.008 & 3.0$\pm$2.6 & 3.37$\pm$0.02 & 18.2$\pm$0.4 & 123$\pm$8 \\
    $K{=}2$, $\eta{=}0.01$ & 0.962$\pm$0.001 & 0 & 3.12$\pm$0.04 & 14.5$\pm$0.3 & 63$\pm$4 \\
    $K{=}5$, $\eta{=}0.01$ & 0.965$\pm$0.001 & 0 & 2.97$\pm$0.02 & 13.5$\pm$0.2 & 58$\pm$4 \\
    $K{=}5$, $\eta{=}0.02$ & 0.961$\pm$0.001 & 0 & 2.94$\pm$0.02 & 13.5$\pm$0.3 & 57$\pm$5 \\
    $K{=}10$, $\eta{=}0.005$ & 0.967$\pm$0.001 & 0 & 2.95$\pm$0.02 & 13.5$\pm$0.4 & 59$\pm$4 \\
    $K{=}5$,\,$\eta{=}0.01$,\,$H{=}8$ & 0.959$\pm$0.001 & 0 & 3.20$\pm$0.03 & 15.7$\pm$0.2 & 82$\pm$6 \\
    $K{=}5$,\,$\eta{=}0.01$,\,last & 0.956$\pm$0.001 & 0 & 3.11$\pm$0.09 & 14.0$\pm$0.3 & 62$\pm$3 \\
    \bottomrule
  \end{tabular}
  \caption{\textbf{\Cref{tab:refine} with standard deviations over seeds} (MPJPE and global MPJPE in mm).
  With $K = 5$ and $\eta = 0.01$ the refinement takes 0.16~s per control step for all 108 clips at once on one
  RTX~5090.}
  \label{tab:refine-seeds}
\end{table}

\section{Refinement in training}
\label{sec:supp-pods}

\begin{table}[h]
  \centering
  \footnotesize
  \setlength{\tabcolsep}{3pt}
  \begin{tabular}{@{}lrrrrr@{}}
    \toprule
    policy update & s/it & 0.83 & 0.90 & best (failed) & MPJPE \\
    \midrule
    BPTT (the recipe) & \best{0.14} & \best{6.4} & \best{8.1} & \best{0.941} (0) & \best{18.7} \\
    BPTT + refined first actions & 0.19 & 9.5 & 11.9 & 0.924 (2) & 19.7 \\
    $\nabla$Mimic's action targets & 0.16 & 15.0 & -- & 0.859 (4) & 30.8 \\
    \bottomrule
  \end{tabular}
  \caption{\textbf{Refinement in training.} One 15-minute run each on \msh: seconds per iteration, minutes to held-out
  returns of 0.83 and 0.90, best held-out return (failed clips) and final MPJPE (mm).}
  \label{tab:pods}
\end{table}

\Cref{tab:pods} compares the recipe's policy update, backpropagation through the imagined window (BPTT), with two
updates that train the policy on actions refined through the world model. \emph{BPTT + refined first actions} keeps the
BPTT step and adds a regression term. Each policy update refines the first action at a quarter of its starts by
\cref{eq:refine} ($K = 2$), with the live policy and the bootstrap critic, and pulls the policy's action toward the
refined one, with the gradient of this term rescaled to the BPTT gradient's norm. The refinement found a better action
for nearly every start. Per iteration, the training return was level with the recipe's or ahead, but the held-out
return was behind at every checkpoint, and the extra rollouts cost 38\% more time per iteration.

\emph{$\nabla$Mimic's action targets}~\cite{lu2026nablamimic} replace the BPTT step with regression. The backward pass
stops at every imagined step's action $e_t$ instead of reaching the policy's parameters. Each action, moved a fixed
distance $\delta$ along its normalized gradient $\partial \hat J / \partial e_t$, becomes a target, and the policy takes
eight regression steps toward the targets. We keep $\nabla$Mimic's settings except for the optimizer (Muon instead of
Adam), the minibatch and $\delta$, which is in excitations instead of joint-setpoint radians. The imagined return rises
faster at first but stalls from iteration 1000 to 2000 and ends at 0.80, against the recipe's 0.94. On held-out clips
it leads only while every clip still fails. The world-model loss stays at 0.25--0.38, where the recipe's falls to 0.04,
and at this learning rate the regression steps overshoot their targets. This run is a single seed, but its gap to the
recipe is far larger than the recipe's spread over seeds.

These two runs suggest four reasons why regression onto refined actions helps $\nabla$Mimic and not us. (i)
$\nabla$Mimic amortizes an expensive gradient, a simulated window and its adjoint, over several regression steps. An
imagined window is cheap, and the BPTT step already uses all of it, so the first variant only adds cost. (ii) With
exact gradients, a target is a real improvement. With a learned model, it is partly the model's error. A policy moved
along that error produces data the model fits poorly, as the second run's world-model loss shows, and the next targets
get worse. Test-time refinement (\cref{sec:refine}) leaves the policy, and so the data, unchanged. (iii) Targets of
fixed size move the policy by the same amount however small the gradient gets: the second run's imagined return fell
over its last 1000 iterations, whereas $\nabla$Mimic anneals $\delta$ and the learning rate. The overshooting regression
steps confound (ii) and (iii), and a smaller step or Adam may behave differently. (iv) The refined action depends on the
root's horizontal error relative to the reference, which our policy does not observe (\cref{sec:supp-lafan}).
Regression onto it can then at best fit its mean given the observation. This would explain why the first variant does
not help, though not why it falls behind the recipe.

\section{Muscle activations and EMG}
\label{sec:supp-emg}

\begin{table*}[t]
  \centering
  \footnotesize
  \setlength{\tabcolsep}{4pt}
  \begin{tabular}{@{}lrrrrrrrrr@{}}
    \toprule
    & strides & \multicolumn{3}{c}{Gait120} & \multicolumn{2}{c}{Lencioni} & mean & activation & held-out \\
    \cmidrule(lr){3-5} \cmidrule(lr){6-7}
    & & $r$ & best lag & $r$, best lag & $r$ & $r$, best lag & activation & $> 0.5$ & return (failed) \\
    \midrule
    ours, \msh (held-out clips, 3 seeds) & 88 & 0.13 & 13\% & 0.55 & $-$0.07 & 0.50 & 0.27 & 22\% & \\
    ours, \mfb (training clips, 3 seeds) & 179 & 0.10 & 14\% & 0.48 & $-$0.06 & 0.41 & 0.44 & 43\% & \\
    MuscleMimic, \mfb, sampled actions & 56 & 0.09 & 14\% & 0.46 & $-$0.04 & 0.30 & 0.49 & 52\% & \\
    MuscleMimic, \mfb, deterministic & 69 & 0.11 & 13\% & 0.46 & $-$0.02 & 0.41 & 0.42 & 42\% & \\
    \midrule
        ours, \msh, one run & 29 & 0.17 & 12\% & 0.54 & $-$0.02 & 0.51 & 0.26 & 20\% & 0.935 (3) \\
    \quad + effort penalty 0.1 & 28 & 0.12 & 12\% & 0.58 & $-$0.12 & 0.58 & 0.077 & 3\% & 0.882 (11) \\
    \quad + effort penalty 0.5 & 26 & 0.25 & 11\% & 0.63 & $-$0.02 & 0.63 & 0.035 & 1\% & 0.823 (22) \\
    \bottomrule
  \end{tabular}
  \caption{\textbf{Muscle activations against walking EMG.} $r$: Pearson correlation of the activation and EMG cycle
  means, averaged over 11 (Gait120) or 8 (Lencioni \etal) muscles, unshifted and at the setting's best common lag
  (Lencioni \etal: 15--18\%). Effort rows: held-out return (failed clips) on the 108 held-out KIT clips. MuscleMimic's
  tracker never activates the vastus lateralis and the gluteus maximus, which are left out of its means.}
  \label{tab:emg}
\end{table*}

\begin{figure*}[t]
  \centering
  \includegraphics[width=\linewidth]{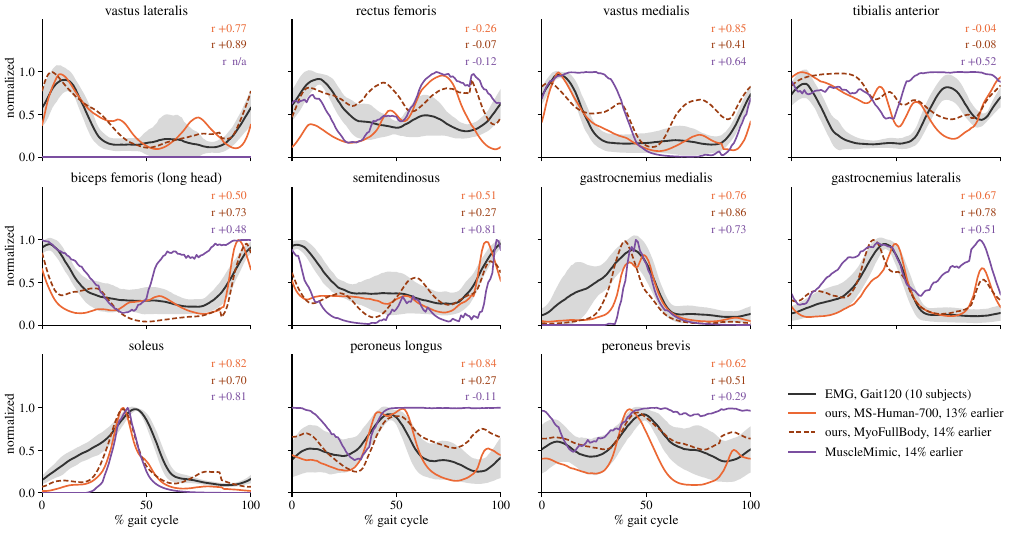}
  \caption{\textbf{Activations against EMG over the gait cycle.} Eleven right-leg muscles from a right heel strike:
  Gait120's EMG (mean and standard deviation over subjects) and the activations of our policies and MuscleMimic's
  tracker, each normalized to its peak and shifted by its best common lag (about 150~ms), with its correlation at that
  lag.}
  \label{fig:emg}
\end{figure*}

We compare the policies' muscle activations with walking EMG under one protocol. Our three 20-minute 50~Hz \msh
policies (\cref{tab:accuracy}) walk the 33 straight walking clips among KIT's held-out clips. Our three 20-minute \mfb
policies and MuscleMimic's released tracker, run with its own code and simulator (\cref{sec:supp-musclemimic}), walk
the same 60 straight walking clips of the training split. A gait cycle starts at a right heel strike, found in the
simulated kinematics as a maximum of the heel's distance ahead of the pelvis~\cite{zeni2008gait}. We average each muscle's activation over the strides at
0.8--1.6~m/s. The references are the right-leg EMG of 11 muscles from Gait120~\cite{boo2025gait120} and the
dominant-leg EMG of 8 muscles from Lencioni \etal~\cite{lencioni2019data}. The two references correlate at 0.87, and
the simulated knee angle agrees with each dataset's at 0.90--0.98, so the gait cycles are aligned. \Cref{tab:emg}
gives the correlations unshifted and at each setting's best common lag.

Under this protocol no tracker's activations resemble the EMG. For our policies on both models and for MuscleMimic's
tracker alike, the correlation is about 0.1 with Gait120 and about 0 with Lencioni \etal (\cref{tab:emg,fig:emg}),
whereas MS-Emulator's policies, regularized with EMG, reach 0.97 under its own protocol~\cite{wei2026msemulator}. Our
policies and MuscleMimic's tracker all co-contract, with a mean activation of 0.27 on \msh and 0.42--0.49 on \mfb. Much of the mismatch is a delay
common to all of them: shifted earlier by about 150~ms, each tracker's activations correlate with the EMG at
0.30--0.55. MuscleMimic's tracker needs the same shift as ours, which suggests that the delay comes from the simulated
muscles rather than from either training method. A likely cause is that MuJoCo's muscle force follows the activation
without the delay by which real muscle force lags the EMG. The shift is chosen after the fact and flatters every
setting, so the unshifted correlations remain the measure.

An effort penalty on the mean squared activation in the training reward lowers the co-contraction 3 and 7 times at
weights 0.1 and 0.5, but it costs tracking and moves the correlation little (\cref{tab:emg}). With a tracking reward,
with or without the penalty, none of these trackers produces activations like EMG. The co-contraction
comes from the policies themselves, not from the exploration noise of \cref{sec:recipe}, which the deployed policy
never executes.

\section{LAFAN1 on both whole-body musculoskeletal models}
\label{sec:supp-lafan}

\begin{table*}[t]
  \centering
  \footnotesize
  \begin{tabular}{@{}lrrrrr@{}}
    \toprule
    & & \multicolumn{2}{c}{MuscleMimic's goal$^*$} & \multicolumn{2}{c}{relative root} \\
    \cmidrule(lr){3-4}\cmidrule(l){5-6}
    held-out clip (subject 5) & length (s) & policy & refined & policy & refined \\
    \midrule
    aiming 2 & 306 & 10.5 & 287.2 & 49.1 & \best{306} \\
    dance 2 & 226 & 16.2 & 138.3 & 182.8 & \best{226} \\
    fight 1 & 245 & 8.4 & \best{245} & 12.4 & \best{245} \\
    jumps 1 & 244 & 7.6 & \best{244} & 14.4 & 182.9 \\
    run 1 & 238 & 7.1 & 22.8 & 9.9 & 25.3 \\
    walk 1 & 261 & 25.3 & \best{261} & \best{261} & \best{261} \\
    walk 3 & 247 & 9.0 & \best{247} & \best{247} & \best{247} \\
    \midrule
    seconds tracked (of 1766) & & 84 & 1445 & 776 $\pm$ 3 & 1493 $\pm$ 86 \\
    clips to their end & & 0 & 4 & 2.5 $\pm$ 0.7 & 5.5 $\pm$ 0.7 \\
    normalized return & & 0.039 & 0.667 & 0.357 $\pm$ 0.008 & 0.722 $\pm$ 0.042 \\
    \midrule
    \multicolumn{6}{@{}l}{\emph{the 173 consecutive 10~s windows of the same clips}} \\
    normalized return & & 0.672 & 0.818 & 0.751 $\pm$ 0.016 & 0.861 $\pm$ 0.001 \\
    failed windows & & 56 & 2 & 15.0 $\pm$ 4.2 & 0.5 $\pm$ 0.7 \\
    MPJPE (mm) & & 58.5 & 43.9 & 51.9 $\pm$ 0.4 & 34.9 $\pm$ 1.0 \\
    joint MAE (\dg) & & 9.27 & 8.39 & 8.04 $\pm$ 0.48 & 6.85 $\pm$ 0.30 \\
    global MPJPE (mm) & & 356 & 166 & 295 $\pm$ 6 & 151 $\pm$ 3 \\
    \bottomrule
  \end{tabular}
  \caption{\textbf{LAFAN1's held-out clips on \msh with a relative-root goal.} One hour of the 50~Hz recipe with
  MuscleMimic's goal ($^*$one run, RTX~5090) or the relative-root goal (mean $\pm$ standard deviation over seeds,
  RTX~4090), by the policy alone and refined ($K = 5$). Top: seconds tracked before a termination (bold: to the clip's
  end in every seed). Bottom: the 173 consecutive 10~s windows of these clips, where repeated evaluations of one
  checkpoint differ by up to 9 failed windows.}
  \label{tab:lafan-relroot}
\end{table*}

We retarget the 77 BVH clips of LAFAN1 other than the fall, push and obstacle clips to \mfb with GMR's inverse
kinematics, with the bone scales and offsets of the IK configuration fitted to \mfb's joint centres (median IK error
1.0--1.4\cm per clip). The clips go from \mfb to \msh as the KIT clips do. Subjects 1--4 train, and subject 5, 7 clips
of 1.5--5 minutes, is held out, each clip evaluated once from its first frame.

\begin{figure*}[t]
  \centering
  \includegraphics[width=\linewidth]{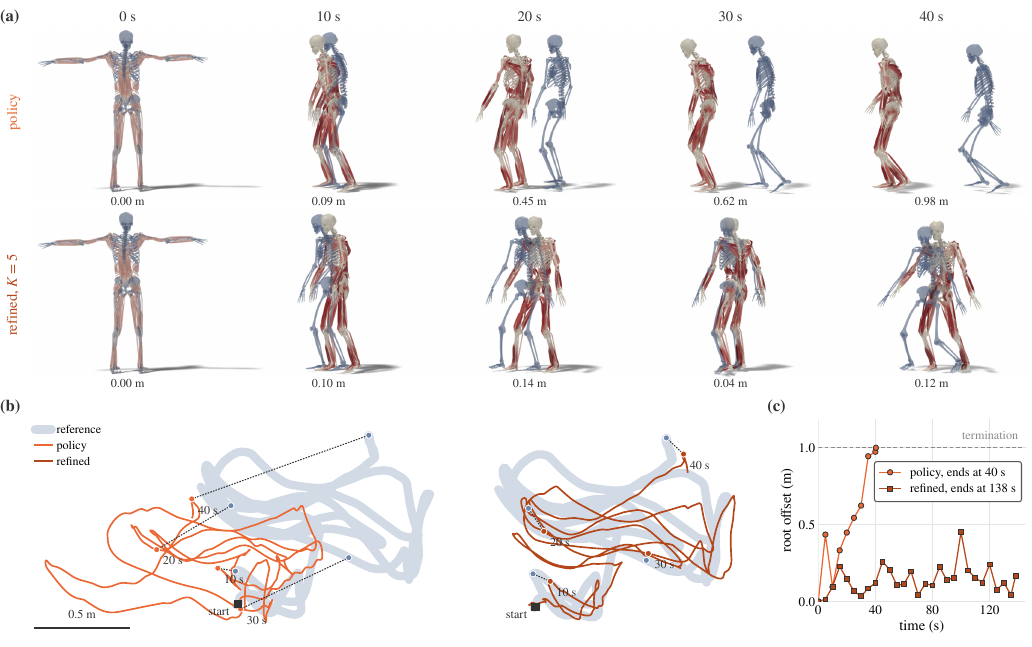}
  \caption{\textbf{Test-time refinement on a long LAFAN1 clip.} Held-out dance 2 on \mfb, the 20-minute policy of
  \cref{tab:lafan-refine} alone and refined ($K = 5$). (a)~Body and reference (blue) at their real horizontal distance.
  (b)~Root paths over the first 40~s from above, offsets at 10--40~s dotted. (c)~Horizontal root distance from the
  reference: the policy is 1~m off at 40~s, and refined the root stays within 0.2~m for 83\% of the time until a posture
  termination at 138~s.}
  \label{fig:lafan-qual}
\end{figure*}

On \msh, the 100~Hz recipe trained for 1.7 hours fails every held-out clip at every checkpoint, after 7--63~s, always
by the root drifting 1~m from the reference in the horizontal plane while the posture is tracked. The same checkpoint
scores 0.80 on KIT's held-out clips, and a KIT-trained policy fails every LAFAN1 clip after 1.2~s. On \mfb, the 50~Hz
recipe fails every held-out clip after 20 minutes of training and after an hour. Without the drift termination 6 and 4
of the 7 clips still fail, so on \mfb the posture terminations end most clips too. \Cref{tab:lafan-refine} adds the
test-time refinement of \cref{sec:refine-method} to these policies, and \cref{fig:lafan-qual} shows one clip with and
without it. Refined with $K = 5$, the one-hour policy tracks 5 of the 7 clips to their end and the 20-minute policy 2.
A single gradient with a line search over the step size reaches the imagined gain of $K = 5$ at 40\% of its cost but
keeps little of the real gain, while on KIT's 5~s clips it keeps 70\% of the MPJPE gain. A likely reason is that on
long clips the sequential small steps, each with a fresh gradient, stay where the model is reliable, and a single large
step does not.

\begin{table}[h]
  \centering
  \footnotesize
  \setlength{\tabcolsep}{2pt}
  \begin{tabular}{@{}lrrrrrrr@{}}
    \toprule
    & & \multicolumn{4}{c}{20 minutes} & \multicolumn{2}{c}{1 hour} \\
    \cmidrule(lr){3-6}\cmidrule(l){7-8}
    clip (subject 5) & length & policy & $K{=}2$ & $K{=}5$ & search & policy & $K{=}5$ \\
    \midrule
    aiming 2 & 306 & 6.7 & 252.4 & \best{305.7} & 79.1 & 29.3 & \best{305.7} \\
    dance 2 & 226 & 44.9 & 138.2 & 138.0 & 89.8 & 49.5 & 138.4 \\
    fight 1 & 245 & 7.9 & \best{244.8} & 216.5 & 12.7 & 21.7 & \best{244.8} \\
    jumps 1 & 244 & 8.0 & 65.4 & 190.1 & 34.0 & 9.9 & 239.9 \\
    run 1 & 238 & 10.3 & 10.3 & 8.9 & 8.0 & 7.7 & \best{237.8} \\
    walk 1 & 261 & 31.6 & 165.7 & 165.7 & 119.2 & 33.6 & \best{261.3} \\
    walk 3 & 247 & 10.7 & 41.1 & \best{246.6} & 117.3 & 25.8 & \best{246.6} \\
    \midrule
    normalized return & & 0.051 & 0.348 & 0.510 & 0.182 & 0.077 & 0.734 \\
    failed clips & & 7 & 6 & 5 & 7 & 7 & 2 \\
    global MPJPE (mm) & & 436 & 305 & 250 & 247 & 367 & 246 \\
    mean imagined gain & & 0 & 0.033 & 0.050 & 0.048 & 0 & 0.037 \\
    \bottomrule
  \end{tabular}
  \caption{\textbf{LAFAN1's held-out clips on \mfb.} Seconds tracked before a termination (bold: to the clip's end) by the
  50~Hz policy after 20 minutes, alone, refined with $K = 2$ or 5 steps, or refined by one gradient with a parallel line
  search over eight step sizes (0.0025--0.32), and by a second run's policy after one hour, alone and refined with
  $K = 5$. Global MPJPE over the tracked time.}
  \label{tab:lafan-refine}
\end{table}

\begin{table}[t]
  \centering
  \footnotesize
  \setlength{\tabcolsep}{3pt}
  \begin{tabular}{@{}lrrr@{}}
    \toprule
    trained on & 972 clips & 864 clips & 864 clips \\
    evaluated on & original 108 & original 108 & new 108 \\
    \midrule
    minutes to 0.83 & 6.3$\pm$0.1 & 5.8$\pm$0.3 & 6.2$\pm$0.0 \\
    normalized return & 0.936$\pm$0.008 & 0.940$\pm$0.003 & 0.943$\pm$0.004 \\
    failed clips & 3.0$\pm$2.6 & 0.7$\pm$1.2 & 0.7$\pm$1.2 \\
    joint MAE (\dg) & 3.37$\pm$0.02 & 3.53$\pm$0.08 & 3.43$\pm$0.05 \\
    MPJPE (mm) & 18.2$\pm$0.4 & 18.8$\pm$0.2 & 18.1$\pm$0.2 \\
    global MPJPE (mm) & 123$\pm$8 & 137$\pm$5 & 118$\pm$9 \\
    root orientation (mrad) & 14.4$\pm$0.6 & 15.5$\pm$0.5 & 15.4$\pm$0.3 \\
    \bottomrule
  \end{tabular}
  \caption{\textbf{A second held-out split.} The 20-minute 50~Hz recipe trained on all 972 training clips (the runs of
  \cref{tab:accuracy}) or on 864 of them, evaluated on the original 108 held-out clips or on the 108 left out, mean
  $\pm$ standard deviation over three seeds.}
  \label{tab:resplit}
\end{table}

\paragraph{Relative-root goal}
On \msh, the 50~Hz recipe trained for one hour also fails every held-out clip, each after 7--25~s by the drift
termination (\cref{tab:lafan-relroot}). The policy cannot see the drift: it observes the root's height but not its
horizontal position, and the goal describes only the reference. Only the reward, the terminations and the critic see
the root error. The relative-root goal adds, at each goal frame $f_k$, the reference root's position and orientation in
the body's root frame, $R^\top(p^{\mathrm{ref}}_{f_k} - p)$ and the 6D representation of $R^\top R^{\mathrm{ref}}_{f_k}$,
with $p$ and $R$ the body root's position and orientation. Imagination computes it from the imagined states and passes
its gradient to them. The official SuperMimic code~\cite{supermimic2026code} gives its actor the relative orientation
and its world model and critic the relative position as well; here the policy gets both. We train two seeds with it for one hour on an RTX~4090, with 22\% fewer iterations than the run
with MuscleMimic's goal on an RTX~5090, and evaluate all three runs on the same machine.

Averaged over seeds (\cref{tab:lafan-relroot}), the relative-root policy alone tracks $2.5 \pm 0.7$ of the seven clips
to their end, where the policy with MuscleMimic's goal tracks none, and it fails fewer of the 10~s windows at a lower
MPJPE. Refined, the relative-root runs track $5.5 \pm 0.7$ clips to their end and the run with MuscleMimic's goal 4.
The new goal thus adds a clip or two and lowers the error, but it does not remove the drift: 8 of the policy's 9
failures across seeds are still the drift termination. The running clip fails within 26~s in every column. \Cref{fig:lafan-rows} shows five of the clips minutes in, refined.

\newsavebox{\lafanrowbox}
\newcommand{\lafanrow}[2]{\sbox{\lafanrowbox}{\includegraphics[width=0.962\linewidth]{fig/lafan_rows/#1.jpg}}%
  \makebox[0.038\linewidth][l]{\rotatebox{90}{\parbox{\ht\lafanrowbox}{\centering\scriptsize #2}}}\usebox{\lafanrowbox}\par}
\begin{figure*}[p]
  \centering
  \lafanrow{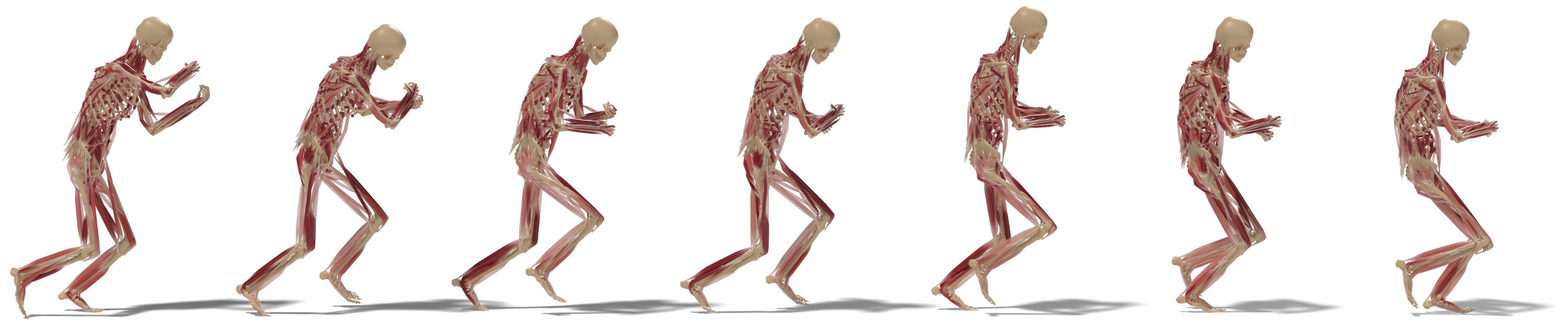}{aiming 2}\vspace{3pt}
  \lafanrow{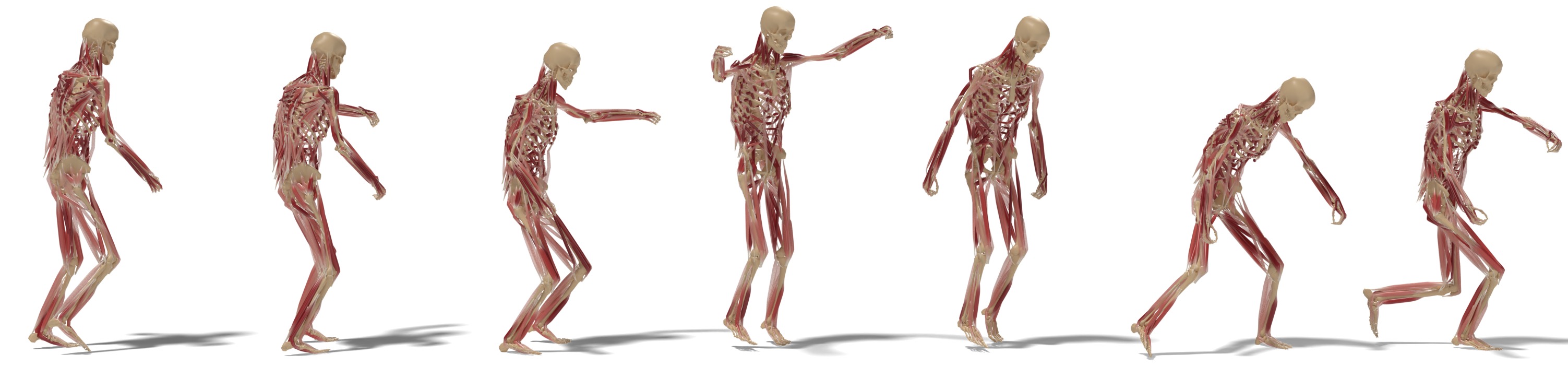}{dance 2}\vspace{3pt}
  \lafanrow{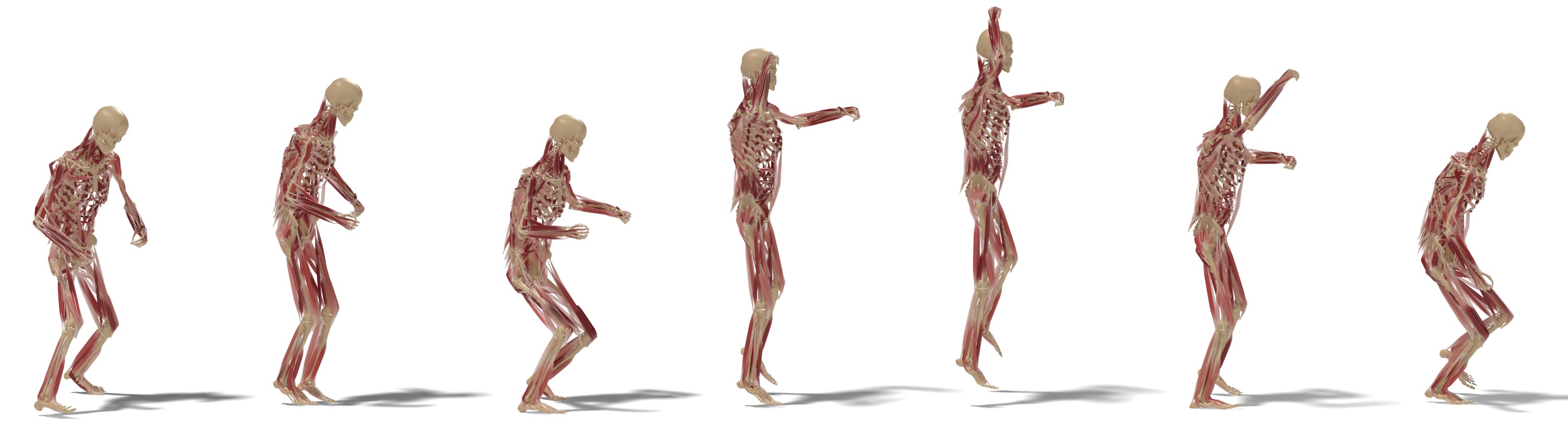}{fight 1}\vspace{3pt}
  \lafanrow{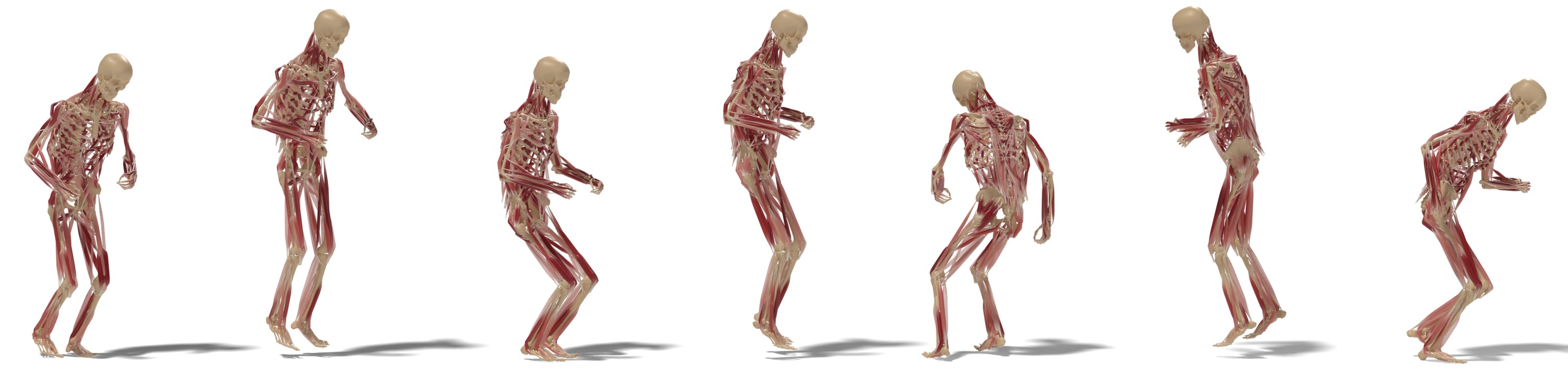}{jumps 1}\vspace{3pt}
  \lafanrow{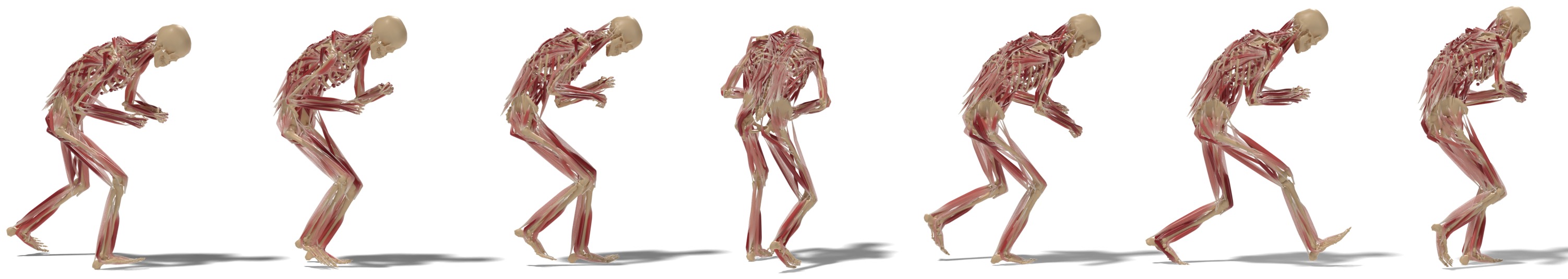}{walk 3}
  \caption{\textbf{Five held-out LAFAN1 clips on \msh, minutes in}, refined ($K = 5$) from the relative-root policy of
  \cref{tab:lafan-relroot}, a few seconds per row. Darker muscles are more activated.}
  \label{fig:lafan-rows}
\end{figure*}

\section{A held-out split not used for any choice}
\label{sec:supp-resplit}

Every choice in the recipe was made on the 108 held-out clips of \cref{tab:accuracy}. To test whether this flatters the recipe, we drew 108 of the 972 training clips at random as a second held-out set. We
trained the final 50~Hz recipe on the other 864 for 20 minutes with three seeds and evaluated it on both held-out sets
(\cref{tab:resplit}). On the new clips every seed reaches 0.83 at iteration 2500, as the runs on all 972 clips do on
the original ones, and ends as high. On the original clips, training on 864 instead of 972 clips reaches 0.83 no later
and costs no return, but adds 0.16\dg of joint MAE and 0.6\mm of MPJPE.

\end{document}